%% file: main.tex
\documentclass{article}

\usepackage{microtype}
\usepackage{graphicx}
\usepackage{booktabs}
\usepackage{hyperref}

\IfFileExists{icml2026.sty}{\usepackage{icml2026}}{\usepackage{icml_fallback}}
\hypersetup{hidelinks,
  pdftitle={Collective Dynamics of Self-Reinforcement: How Cross-Task Coupling Drives Policy Collapse in Unsupervised RLVR},
  pdfauthor={Peize Ding}}

\usepackage{amsmath,amssymb,mathtools,amsthm}
\usepackage[capitalize,noabbrev]{cleveref}
\theoremstyle{plain}
\newtheorem{proposition}{Proposition}

\theoremstyle{remark}

\icmltitlerunning{Collective Dynamics of Self-Reinforcement in Unsupervised RLVR}

\begin{document}

\twocolumn[
\icmltitle{Collective Dynamics of Self-Reinforcement:\\
How Cross-Task Coupling Drives Policy Collapse in Unsupervised RLVR}

\icmlsetsymbol{equal}{*}
\begin{icmlauthorlist}
\icmlauthor{Peize Ding}{cu}
\end{icmlauthorlist}
\icmlaffiliation{cu}{Columbia University, New York, NY, USA}
\icmlcorrespondingauthor{Peize Ding}{pd2714@columbia.edu}
\icmlkeywords{reinforcement learning, RLVR, label-free RL, test-time RL, collapse, statistical physics}

\vskip 0.3in
]

\printAffiliationsAndNotice{}

\input{sec_abstract}
\input{sec_intro}
\input{sec_theory}
\input{sec_experiments}

\input{sec_discussion}

\bibliography{references}
\IfFileExists{icml2026.bst}{\bibliographystyle{icml2026}}{\bibliographystyle{plainnat}}

\newpage
\appendix
\crefalias{section}{appendix}
\crefalias{subsection}{appendix}
\onecolumn
\input{sec_appendix}

\end{document}

%% file: sec_abstract.tex
\begin{abstract}
Label-free reinforcement learning with verifiable rewards rewards a language model for agreeing with its own majority vote. It improves reasoning at first but eventually collapses: the self-assessed reward goes to one while accuracy falls, often until the model gives the same answer to every problem. Empirically, the collapse appears only when training uses many problems.
We explain this dataset-size dependence as a collective instability.
In a solvable many-problem model in which problems share parameters, a population in which different problems prefer different answers amplifies any shared perturbation for as long as the population-averaged slope $\langle F'\rangle$ of the single-problem self-reinforcement drift $F$ is positive. This holds while problems are still uncertain about their answers.
Per-problem sharpening, slowed by $1/N$ for $N$ training problems, races against the shared mode, which grows at a rate $\propto\lambda^2$ independent of $N$, where $\lambda$ is the strength of the shared coupling. This yields a critical number of problems $N_c\simeq\ln(1/\delta)/(\lambda^2G)$, where $\delta$ is the noise that seeds the shared mode and $G=\ln[F^\star/F(a)]$ decreases with the initial confidence $a$.
Stochastic simulations of the model confirm the predictions: a sharp transition, $N_c\propto\lambda^{-2}$ up to a logarithmic correction, a single critical amplification $\lambda^2NG$ across couplings, confidence profiles and training variants, a merely logarithmic dependence on batch size, and no collapse for confident initial models.
A kernel version of the theory identifies a second, protective coupling (generalization of sharpening) and a marginal-entropy bonus that stabilizes the uniform collective mode.
In label-free RL of a small Transformer, removing the shared direction prevents collapse entirely, restoring it causes collapse, and collapse grows with training-set size and base-model entropy.
\end{abstract}

%% file: sec_intro.tex
\section{Introduction}
\label{sec:intro}

Reinforcement learning with verifiable rewards (RLVR) is now a central ingredient in training reasoning models \citep{tulu3,deepseekmath,deepseekr1,qwen3}.
The recipe is simple. Sample several solutions to a problem, reward those whose final answer matches the reference, and raise the probability of rewarded solutions with a policy gradient such as group relative policy optimization (GRPO; \citealp{deepseekmath}).
Its bottleneck is equally simple: every training problem needs a trusted reference answer.
Such answers are expensive to produce, and they become harder to obtain as models approach expert ability.

This has motivated \emph{label-free} or \emph{unsupervised} RLVR, in which the model rewards itself.
The most common choice is majority voting \citep{ttrl,shafayat2025srt}. The answer that appears most often among the $K$ samples is treated as a pseudo-label, and samples that agree with it are rewarded.
Related intrinsic signals include self-certainty and entropy \citep{intuitor,agarwal2025entmin,rent,empo}.
These methods often help early in training, but with prolonged training they \emph{collapse}.
The self-assessed reward keeps rising while true accuracy falls \citep{shafayat2025srt,he2026urlvr}, and in the extreme the model emits the same answer for every problem \citep{shafayat2025srt}.

A striking feature of this collapse is its dependence on the \emph{number of training problems}.
In a systematic study, \citet{he2026urlvr} found that training on $\le128$ problems was stable, whereas ``larger datasets ($\ge512$) consistently exhibit reward hacking''.
The same algorithm is safe with few problems and fails with many.
This is the signature of a \emph{collective} phenomenon, and it is not explained by existing theory.
The theory of intrinsic rewards is single-problem: each problem is sharpened toward its initial majority answer \citep{huang2025sharpening,he2026urlvr}.
Such sharpening helps when the initial majority is correct and hurts when it is wrong, but it says nothing about why \emph{all} problems should be dragged toward a common answer, or why that should depend on how many there are.

\paragraph{This paper.}
We argue that dataset-size-dependent collapse is an instability of a \emph{shared direction} in parameter space.
Problems are coupled because they share network parameters \citep{jacot2018ntk,ren2025learning,fan2026sharpening}.
Any parameter that moves all answers the same way (a format, a preferred value, an output bias) collects the majority-vote signal from every problem at once, whereas each problem-specific parameter collects only its own share.
Our contributions are:
\begin{itemize}\setlength\itemsep{1pt}
\item \textbf{Mechanism.} In a solvable many-problem model, a shared perturbation of the state in which different problems prefer different answers grows, at each instant along the sharpening trajectory, exactly when the average slope $\langle F'\rangle$ of the single-problem self-reinforcement drift $F$ is positive. For $K=3$ samples per problem this holds when problems are still uncertain, with answer logits $|z|<\ln(2+\sqrt3)\approx1.32$, a Curie--Weiss-type paramagnetic instability (\cref{sec:theory}).
\item \textbf{A critical dataset size.} Local sharpening, slowed by $1/N$ for $N$ training problems, races against the shared mode, which grows at a rate $\propto\lambda^2$ independent of $N$, where $\lambda$ is the strength of the shared coupling. The result is a closed-form critical size $N_c\simeq\ln(1/\delta)/(\lambda^2\ln[F^\star/F(a)])$, where $\delta$ is the noise that seeds the shared mode, $a$ is the initial confidence and $F^\star$ is the drift at the edge of the uncertain range. It makes four falsifiable predictions.
\item \textbf{Verification.} Stochastic simulations with actual sampling, majority voting and minibatch stochastic gradient descent (SGD) confirm all four predictions. The transition is sharp; $N_c\propto\lambda^{-2}$ up to a logarithmic correction, with a single critical value $\Lambda_c\approx2.6$ of the amplification $\lambda^2N\ln[F^\star/F(a)]$ that also captures the dependence on initial confidence; batch size enters only logarithmically; and confident initial models never collapse, even under much longer training. The same law holds for GRPO, larger groups, many answers, heterogeneous confidence and Adam, the last with a larger constant (\cref{sec:sims,sec:robust}).
\item \textbf{A kernel version for networks.} Linearizing a network gives a version of the theory in terms of the empirical neural tangent kernel (NTK). It reveals a second, protective coupling, the generalization of sharpening across problems, under which adding data can stop causing collapse (\cref{sec:kernel}).
\item \textbf{A small Transformer.} In label-free RL, removing the shared direction by an antisymmetric readout prevents collapse entirely, adding it back causes collapse, and collapse grows with training-set size and base-model entropy. Predictions from the initial kernel rank runs by their risk of collapse, but they are poorly calibrated: they underestimate collapse and agree with the outcome no more often than always predicting collapse. Measuring the kernel during training identifies one source of this error, the erosion of the protective coupling by feature learning (\cref{sec:nn}).
\item \textbf{Stabilizing the uniform mode, and its limit.} A batch-level marginal-entropy bonus with weight $\beta$ damps the uniform shared mode. Its critical weight $\beta_c\le\tfrac12$ is derived from the theory and confirmed in simulation and in the Transformer. In the network, however, the instability then moves to the next collective kernel mode, so removing the imbalance of the answer marginal does not prevent a task-unrelated collective collapse, and accuracy is not preserved.
\end{itemize}
After reviewing related work, we analyze a single problem in isolation (\cref{sec:single}) and then couple many problems through a shared parameter (\cref{sec:theory}).
We test the resulting predictions in stochastic simulations (\cref{sec:sims}, with further robustness checks in \cref{sec:robust}), carry them over to a small Transformer (\cref{sec:nn}), and close with implications and limitations (\cref{sec:discussion}).

\section{Related Work}
\label{sec:related}

\paragraph{Label-free RL for reasoning.}
Test-time reinforcement learning (TTRL; \citealp{ttrl}) uses majority-vote pseudo-rewards on unlabeled test problems.
Self-training with majority votes \citep{shafayat2025srt}, self-certainty \citep{intuitor}, entropy minimization \citep{agarwal2025entmin,rent} and semantic-entropy objectives \citep{empo} are close relatives. Even random or spurious rewards can help some base models \citep{spurious}.
Several works report eventual reward hacking or collapse and propose remedies such as cross-view rewarding \citep{corewarding}.
\citet{he2026urlvr} study unsupervised RLVR systematically. They show that intrinsic methods converge toward sharpening the model's initial distribution, that collapse depends on dataset size, and that base models with higher initial entropy fare worse.
We supply a mechanism for the dataset-size dependence and a quantitative criterion.

\paragraph{Sharpening and entropy collapse.}
\citet{huang2025sharpening} formalize self-improvement as sharpening toward high-likelihood responses.
\citet{cui2025entropy} analyze entropy collapse in RL for reasoning models through the covariance between log-probabilities and advantages.
These analyses treat a single prompt, or an aggregate over prompts, without the cross-problem coupling that drives our transition.

\paragraph{Coupling through shared parameters.}
The NTK \citep{jacot2018ntk} and the learning-dynamics view of fine-tuning \citep{ren2025learning} describe how an update on one example moves predictions on another.
Closest to us, \citet{fan2026sharpening} show for \emph{labeled} RLVR that finite-batch updates bias learning toward sampled modes and that collapse propagates through semantic coupling. Their analysis is a one-step linearization for a pair of queries.
We study the collective, many-problem dynamics of \emph{self}-rewarded RL and derive a size threshold.

\paragraph{Statistical physics of self-training and recursive training.}
\citet{takahashi2026selftraining} analyzes iterative self-training of linear classifiers with the replica method.
Model collapse under recursive training on generated data \citep{shumailov2024collapse} is a related but distinct failure that operates across model generations.
Our analysis uses mean-field and linear-stability tools \citep{nishimori2001}. It concerns on-policy sampling with majority-vote policy gradients, where the reward depends on the policy itself.

%% file: sec_theory.tex
\section{Single-Problem Self-Reinforcement}
\label{sec:single}

We start with one problem in isolation. Its sharpening drift $F$, and in particular the slope $F'$, are the only single-problem ingredients that the collective theory of \cref{sec:theory} needs.

\paragraph{Policy, reward and drift.}
We abstract a problem $x$ by the distribution of its \emph{final answers}.
In the binary case the two answers are $+$ and $-$, and the policy is $\pi(+\mid x)=p=\sigma(z)$, $\pi(-\mid x)=1-p$, with logit $z$.
For each problem the learner samples $K$ answers $y_1,\dots,y_K$, forms the majority $\hat a$ (ties broken uniformly at random), and assigns $r_j=\mathbf 1[y_j=\hat a]$ \citep{ttrl}.
The per-sample policy-gradient estimate is $\hat g=\frac1K\sum_j A_j\,\partial_z\log\pi(y_j)$, with $\partial_z\log\pi(+)=1-p$, $\partial_z\log\pi(-)=-p$, and advantage $A_j=r_j$ (REINFORCE; \citealp{williams1992}), $A_j=r_j-\bar r$ (group-mean baseline), or $A_j=(r_j-\bar r)/\mathrm{std}(r)$ (GRPO; \citealp{deepseekmath}).

The estimate $\hat g$ averages over the $K$ answers drawn for a single update, so it is still random: each update draws fresh answers. Its expectation over the draw of $y_1,\dots,y_K\sim\pi$ and the tie-breaking is the mean update on the logit, $F(z)=\mathbb E[\hat g]$.
Because $\hat g$ depends on the samples only through the number $k$ of $+$ answers, $F$ can be computed exactly by summing over $k\sim\mathrm{Bin}(K,p)$ for any $K$ and estimator (\cref{app:drift}).
For REINFORCE it has a simple form. By exchangeability $F=\mathbb E[r_1\,\partial_z\log\pi(y_1)]$, so $F(z)=p(1-p)\big(\Pr[\hat a=+\mid y_1=+]-\Pr[\hat a=-\mid y_1=-]\big)$.
For odd $K=2m+1$ the bracket equals $\Pr[X>m]-\Pr[X<m]$, where $X\sim\mathrm{Bin}(2m,p)$ counts the $+$ answers among the other $K-1$ samples.
Our main experiments use $K=3$. There the bracket is $p^2-(1-p)^2=2p-1$, which gives
\begin{equation}
  F(z)=p(1-p)(2p-1),\qquad p=\sigma(z).
  \label{eq:F}
\end{equation}
A problem trained on its own with step $\eta$ therefore follows, on average, the \emph{sharpening equation}
\begin{equation}
  \dot z=\eta\,F(z),
  \label{eq:sharp}
\end{equation}
where time counts updates.
For every $K\ge3$ and all three estimators, $F$ is odd, positive for $z>0$, and vanishes at $z=0$ and as $|z|\to\infty$. Under \cref{eq:sharp} every problem is therefore pushed toward its current majority answer and then freezes.
This is the single-problem sharpening behaviour characterized by \citet{huang2025sharpening} and \citet{he2026urlvr}.

\paragraph{The unstable window.}
The drift $F$ describes how a problem sharpens on its own. Once problems share parameters, they also feel a common push, and how a problem responds to that push is set by the \emph{slope} $F'(z)$. This slope is the quantity that controls the collective behaviour.
For $K=3$, differentiating \cref{eq:F} and writing $q=p-\tfrac12$ gives $F'(z)=p(1-p)\,(\tfrac12-6q^2)$. Setting $6q^2=\tfrac12$ locates the zero of $F'$, so
\begin{equation}
  F'(z)>0 \iff |z|<z^\star=\ln(2+\sqrt3)\approx1.317,
  \label{eq:window}
\end{equation}
that is, $0.21<p<0.79$. Evaluating \cref{eq:F} at the edge gives $F^\star=F(z^\star)=1/(6\sqrt3)\approx0.096$.
Inside this window a problem's response to a small common push grows with the push. Outside the window it shrinks (\cref{fig:mechanism}a).

So far we have used REINFORCE with $K=3$. We now check how $F$ changes with the estimator and the group size.
For group-mean and GRPO advantages, unanimous groups have zero advantage, so only the $2{:}1$ splits contribute, and they all have the same reward standard deviation. For REINFORCE, unanimous groups also contribute; in fact for $K=3$ they carry the entire mean drift, while the contributions of the $2{:}1$ splits cancel.
Nevertheless, exact enumeration shows that the three estimators have proportional mean drifts at $K=3$: the mean baseline gives exactly $\tfrac23F$ and GRPO exactly $\sqrt2F$ (\cref{app:drift}). \emph{For $K=3$, the choice of estimator changes the time scale but not the shape of $F$}.
For larger $K$ the shapes differ. Each $K$ and estimator then has its own exact drift $F_K$, window edge $z^\star_K$ (where $F_K'=0$) and edge drift $F^\star_K=F_K(z^\star_K)$.
\Cref{tab:window} (Appendix) lists $z^\star_K$, which does not increase with $K$, so larger groups narrow the unstable window.
The analysis below uses $F$ only through the sign of $F'$ and the ratio $F^\star/F$, so it holds for any $K$ and estimator once $F$, $z^\star$ and $F^\star$ are replaced by $F_K$, $z^\star_K$ and $F^\star_K$. We quote $K=3$ values and drop the subscript unless stated otherwise.

\section{Collective Theory}
\label{sec:theory}

We now couple many problems through shared parameters. We introduce a minimal model, show that a balanced population amplifies shared perturbations while its problems are undecided, derive the critical number of problems from a race between two time scales, and extend the result to heterogeneous confidence and many answers. We then give a kernel form of the theory for neural networks and show how to stabilize the uniform collective mode.

\subsection{A minimal many-problem model}
Consider $N$ binary problems whose logits share one parameter:
\begin{equation}
  z_i=u_i+\lambda v,\qquad i=1,\dots,N.
  \label{eq:model}
\end{equation}
$u_i$ is problem-specific, $v$ is a shared direction, and $\lambda$ measures how strongly each problem's logit loads on it.
In a language model, $v$ stands for anything that moves all answers the same way: a format preference, a bias toward short answers, a preferred numerical value, or the output bias of an answer token.
Training uses minibatch SGD with step $\eta$. Each step draws $B$ problems and averages their contributions, $\theta\leftarrow\theta+\frac{\eta}{B}\sum_{i\in\text{batch}}\hat g_i\,\nabla_\theta z_i$, where $\hat g_i$ is the estimate of \cref{sec:single} for problem $i$ and $\theta=(u_1,\dots,u_N,v)$. Since $\partial z_i/\partial u_j=\delta_{ij}$ and $\partial z_i/\partial v=\lambda$, the step moves $u_i$ by $\frac{\eta}{B}\hat g_i$ and $v$ by $\frac{\eta\lambda}{B}\sum_{i\in\text{batch}}\hat g_i$.
A problem is in the batch with probability $B/N$, and $\mathbb E[\hat g_i]=F(z_i)$, so the expected step does not depend on $B$. The batch size affects only the noise.
Replacing the stochastic update by this expected drift, which is accurate for small steps, gives
\begin{align}
  \dot u_i&=\tfrac{\eta}{N}F(z_i), &
  \dot v&=\tfrac{\eta\lambda}{N}\textstyle\sum_jF(z_j),\nonumber\\
  \Rightarrow\ \dot z_i&=\underbrace{\tfrac{\eta}{N}F(z_i)}_{\text{local sharpening}}+\underbrace{\eta\lambda^2\langle F\rangle}_{\text{shared drift}},
  \label{eq:mf}
\end{align}
where $\langle\cdot\rangle$ averages over problems.
We call \cref{eq:mf} the mean-field equations. Because the problems interact only through $v$, which sums all their gradients, the average $\langle F\rangle$ enters exactly. The only approximation is dropping the sampling noise. Without noise a perfectly balanced population would stay balanced forever, so noise re-enters below only as the seed $\delta$ of the shared mode.
The two terms scale differently with $N$. Each problem receives a $1/N$ share of the training signal, but the shared parameter aggregates the signal of all problems.

\subsection{Shared perturbations grow while problems are undecided}
We first ask whether a balanced population, in which different problems prefer different answers, resists a common push.
Suppose the initial preferences are balanced: half the problems lean to $+$ and half to $-$, with $|z_i(0)|=a$.
The constant $a>0$ is the \emph{initial confidence}, the logit magnitude shared by all problems. By symmetry the problems stay at $\pm a(t)$ as they sharpen, with $a(0)=a$.
By oddness of $F$, $\langle F\rangle=0$, so $v\equiv0$ is a solution and each problem sharpens independently.
Perturb it by a common logit shift $w=\lambda\,\delta v$. Linearizing \cref{eq:mf} gives the following (derivation in \cref{app:stability}).

\begin{proposition}[Shared-mode instability]
\label{prop:instab}
To leading order in $1/N$, a common shift evolves as $\dot w=\eta\lambda^2\langle F'(z(t))\rangle\,w$, with the logits $z(t)$ evaluated along the sharpening trajectory.
A small shared perturbation therefore grows at time $t$ if and only if $\langle F'(z(t))\rangle>0$. For a homogeneous population this means $a(t)<z^\star$.
\end{proposition}

\Cref{prop:instab} is a statement about the \emph{instantaneous} growth rate of a perturbation along a time-dependent trajectory, on which every $|z_i|$ keeps increasing. It is not the instability condition of a fixed point, and it is not sufficient for collapse: a shared perturbation that grows for a while can still be too small to flip any answer by the time sharpening has carried the population out of the window. Whether collapse occurs is decided by the \emph{integrated} growth, which we compute in \cref{sec:race}.

The structure is that of the paramagnetic instability of the Curie--Weiss model \citep{nishimori2001}.
$w$ plays the role of a magnetization, the problems are spins coupled only through the mean field, and $\langle F'\rangle$ is the inverse of a susceptibility that changes sign.
Undecided problems ($|z|<z^\star$) collectively amplify \emph{any} shared bias, whether it comes from sampling noise or a slight initial asymmetry.
Confident problems ($|z|>z^\star$) collectively suppress it.

\subsection{A race between sharpening and the shared mode}
\label{sec:race}
\Cref{prop:instab} says when the shared mode grows, but not whether it grows enough to change any answer.
The instability is transient, because local sharpening moves every problem out of the window.
The outcome is decided by which process wins.
Each problem sharpens as in \cref{eq:sharp}, slowed by the factor $1/N$: $\dot a=\frac{\eta}{N}F(a)$. The total log-amplification of the shared mode while the population is inside the window is
\begin{equation}
  \Lambda=\int\!\eta\lambda^2F'(a(t))\,dt=\lambda^2N\,G(a),
  \label{eq:Lambda}
\end{equation}
with $G(a)=\int_a^{z^\star}(F'/F)\,dx=\ln[F^\star/F(a)]$ (\cref{app:stability}).
Local sharpening takes time $\propto N$, and the shared mode grows at an $N$-independent rate $\propto\lambda^2$, so a larger training set gives the shared mode more time to grow.
If the seed of the shared mode has size $\delta$, the minority problems are flipped once $\delta e^{\Lambda}=O(1)$, which gives a critical number of problems
\begin{equation}
  \boxed{N_c\simeq\frac{\ln(1/\delta)}{\lambda^2\,\ln[F^\star/F(a)]}}
  \label{eq:Nc}
\end{equation}
Above $N_c$ every problem converges to the same answer. Accuracy drops to the base rate while the majority-vote reward goes to~1.
For other group sizes and estimators, $G$ is replaced by $G_K(a)=\ln[F^\star_K/F_K(a)]$, built from the exact drift of \cref{sec:single}. \Cref{tab:window} lists $G_K(0.5)$, which also does not increase with $K$. A constant factor in $F$ cancels in the ratio, so $G_3=G$ for all three estimators.

\Cref{eq:Nc} leaves the seed $\delta$ unspecified.
In stochastic training $\delta$ comes from minibatch sampling noise injected into $v$.
Let $\sigma$ be the standard deviation of the single-problem estimate $\hat g_i$, which fluctuates because each update draws fresh answers. Averaging over a batch of $B$ problems gives per-step noise $\eta\lambda^2\sigma/\sqrt B$ on the common shift. With this noise accumulated over the growth time $1/(\eta\lambda^2F')$, we get $\delta^2\sim\eta\lambda^2\sigma^2/(BF')$ (\cref{app:seed}).
The seed enters $N_c$ only through $\ln(1/\delta)=\tfrac12\ln B+\ln(1/\lambda)+\text{const}$, so batch size has only a logarithmic effect. Under the linear scaling rule $\eta\propto B$ it has no effect at all.

\paragraph{Falsifiable predictions.}
Together, these results give four predictions, which we test in \cref{sec:sims}.
(P1)~A sharp transition in $N$ emerges from the dynamics and is not imposed.
(P2)~$N_c\propto\lambda^{-2}$, up to the logarithmic correction above.
(P3)~Batch size enters only logarithmically.
(P4)~In the homogeneous fixed-coupling model, if $a>z^\star$, a small shared perturbation decays at every instant, for every $N$ and every training length. This is a stability property, not a time-scale effect. It is a prediction for this model only: in a network whose kernel changes during training, a small initial $G$ does not guarantee stability (\cref{sec:nn}).
A corollary is that higher-entropy (less confident) initial models have more problems inside the window and therefore collapse at smaller $N$, which matches the empirical observation of \citet{he2026urlvr}.

\subsection{Generalizations}
\label{sec:general}
The analysis so far assumes a homogeneous population of binary problems. We now relax both assumptions.

For arbitrary initial logits, \cref{eq:Lambda} becomes $\Lambda=\lambda^2N\,G[\rho_0]$ (\cref{app:stability}), where
\begin{equation*}
  G[\rho_0]=\sup_s\int_0^s\langle F'(z(s'))\rangle\,ds'
\end{equation*}
and each $z$ follows the sharpening equation \cref{eq:sharp}, $dz/ds=F(z)$ in rescaled time $s=\eta t/N$, from its initial value.
The integrand is the instability index $\langle F'\rangle$ of \cref{prop:instab}, evaluated along the sharpening trajectory.

Next we drop the restriction to two answers.
With $M$ candidate answers, each problem has softmax logits $z_i=u_i+\lambda v\in\mathbb R^M$, and the shared parameter $v\in\mathbb R^M$ is a bias toward some answers over others.
Let $g(z)\in\mathbb R^M$ be the expected single-problem update, the analogue of $F$.
In the symmetric state each problem prefers one answer by a margin $a$, and the preferred answers are spread evenly over the $M$ labels.
A common shift $w=\lambda\,\delta v$ then evolves as $\dot w=\eta\lambda^2\bar Jw$, where $\bar J$ is the Jacobian of $g$ averaged over problems.
For a single problem the Jacobian singles out its preferred answer and couples the components of $w$.
Averaging over problems removes this coupling: every label is preferred by the same number of problems, so $\bar J$ treats all labels alike.
Adding the same constant to all logits leaves a softmax unchanged, so only the $M-1$ zero-sum directions of $w$ matter, and on them $\bar J=\kappa(a)\,I_{M-1}$ (\cref{app:multi}).
The shared mode therefore splits into $M-1$ decoupled directions, each obeying $\dot w=\eta\lambda^2\kappa(a)\,w$.
Each is the binary problem of \cref{prop:instab} with $F'$ replaced by $\kappa$.
Likewise the margin $a$ between the preferred and the other logits sharpens as $\dot a=\frac\eta Nf(a)$. Here $f=g_s-g_{\neq s}$ is the drift of the logit of the preferred answer $s$ minus that of any other answer, and it plays the role of $F$.
The race of \cref{sec:theory} carries over unchanged. The window is $a<a^\star$ with $\kappa(a^\star)=0$, and $\Lambda=\lambda^2N\,G_M(a)$ with $G_M(a)=\int_a^{a^\star}\kappa(x)/f(x)\,dx$.
Both $\kappa$ and $f$ follow from exact enumeration over multinomial count vectors (\cref{app:drift,app:multi}). For $M=2$ they reduce to $F'$ and $F$.

\subsection{Neural networks: a kernel form}
\label{sec:kernel}
The results so far keep the coupling of \cref{eq:model}, a single shared direction with a fixed strength $\lambda$. A real network instead couples problems through all of its parameters.
For a general network, linearizing the parameter dynamics gives $\dot z=\frac{\eta}{N}\mathbf K F(z)$, where $\mathbf K_{ij}=\nabla_\theta z_i\!\cdot\!\nabla_\theta z_j$ is the empirical neural tangent kernel \citep{jacot2018ntk}. $\mathbf K_{ij}$ measures how much a gradient step on problem $j$ moves the logit of problem $i$.
Two collective directions matter. One is the uniform mode $\mathbf 1$, which is the shared direction above. The other is the \emph{sign mode} $s$, with $s_i=\mathrm{sign}\,z_i(0)$ the answer problem $i$ currently prefers.
Moving along $\mathbf 1$ shifts every problem toward the same answer, which is collapse. Moving along $s$ makes every problem more confident in its own answer, which is sharpening applied to all problems at once.
A kernel component $k_s s_is_j$ means that a step sharpening problem $j$ also pushes $z_i$ toward $s_i$, so sharpening one problem sharpens the others. This is how a network generalizes, and it competes with the uniform mode.
We summarize $\mathbf K$ by four numbers.
The \emph{self-kernel} $k_d$, the mean of $\mathbf K_{ii}$, is how much a step on a problem moves its own logit, that is, the effective step size.
The \emph{shared coupling} $k_u$, the mean of $\mathbf K_{ij}$ over $i\ne j$, is how much a step on one problem moves every other logit the same way, whatever their answers. It drives collapse and plays the role of $\lambda^2$.
The \emph{prediction-aligned coupling} $k_s$ is the mean of $s_is_j\mathbf K_{ij}$ over $i\ne j$. It protects against collapse.
The remainder $k_d'=k_d-k_u-k_s$ is the purely problem-specific part of the self-kernel and sets the rate of local sharpening.
Projecting $\mathbf K\approx k_d'\mathbf I+k_u\mathbf 1\mathbf 1^\top+k_s ss^\top$ onto these modes and repeating the race of \cref{sec:race} yields (\cref{app:kernel})
\begin{equation}
  \Lambda(N)=G\,\frac{N\,k_u}{k_d'+N\,k_s},\qquad
  N_c=\frac{\Lambda_c\,k_d'}{G\,k_u-\Lambda_c\,k_s}.
  \label{eq:kernel}
\end{equation}
The projection keeps exactly the parts of $\mathbf K$ that enter the race: its diagonal and its action along $\mathbf 1$ and $s$, since $\mathbf 1^\top\mathbf K\mathbf 1/N\approx k_d'+Nk_u$ and $s^\top\mathbf K s/N\approx k_d'+Nk_s$.
It is exact in the toy model, which is the special case $k_d'=1$, $k_u=\lambda^2$, $k_s=0$. In a network it drops the other collective modes of $\mathbf K$, which we return to in \cref{app:nnremedy}.
When $k_s>0$ the growth of $\Lambda$ with $N$ saturates at $Gk_u/k_s$, and there is a \emph{feature-protected} regime $G\,k_u<\Lambda_c\,k_s$ in which no training-set size causes collapse.
All quantities in \cref{eq:kernel} can be measured on the initial model before any RL step.
This assumes that $\mathbf K$ stays near its initial value, as in the lazy regime. When the network learns features, its gradients $\nabla_\theta z_i$ and hence $\mathbf K$ change, and the prediction can fail. Re-measuring $\mathbf K$ later does not help once the race is decided, and after collapse $k_s=k_u$ because all $s_i$ agree.
In the toy model, a feature coupling $z_i\mathrel{+}=\mu s_iw$ realizes $k_s=\mu^2$ and gives $N_c=1/(\lambda^2G/\Lambda_c-\mu^2)$.

\subsection{Stabilizing the uniform collective mode}
\label{sec:remedy}
Finally, we turn from diagnosing the instability to damping it.
In the minimal model the instability lives entirely in the uniform shared mode, so it can be targeted without slowing per-problem sharpening. A network has further collective modes, and damping the uniform one alone does not protect accuracy there (\cref{sec:nnresults}).
We add to each minibatch objective a marginal-entropy bonus $\beta\,H(\bar p)$ with weight $\beta\ge0$. Here $\bar p=\frac1B\sum_{i\in\text{batch}}p_i$ is the probability of answer $+$ averaged over the batch, and $H$ is the binary entropy
\begin{equation*}
  H(q)=-q\ln q-(1-q)\ln(1-q).
\end{equation*}
The majority-vote reward lowers the entropy of each problem's answer, and the bonus raises the entropy of the answers pooled across problems, in the spirit of mutual-information objectives for unsupervised classification \citep{bridle1991phantom,gomes2010rim}.
When answers are balanced, $\bar p=\tfrac12$ is the maximum of $H$, so the bonus exerts no force and per-problem sharpening is unaffected.
A common shift moves every $p_i$ the same way and pushes $\bar p$ away from $\tfrac12$, and the bonus pushes back.
Near the symmetric state it contributes $-4\beta\langle p(1-p)\rangle^2$ to the growth rate of the shared mode (\cref{app:remedy}), so the shared mode is damped when
\begin{equation}
  \beta>\beta_c(a)=\frac{F'(a)}{4\,[p(1-p)]^2}\le\tfrac12 ,
  \label{eq:beta}
\end{equation}
with $p=\sigma(a)$. The threshold is largest at $a=0$, where $F'(0)=1/8$ and $p(1-p)=\tfrac14$ give exactly $\tfrac12$, so any $\beta>\tfrac12$ damps the uniform shared mode at every confidence.

%% file: sec_experiments.tex
\section{Simulations of the Solvable Model}
\label{sec:sims}

We now test the theory of \cref{sec:theory} in stochastic simulations of the toy model: first the predictions P1--P4, then the extensions of \cref{sec:general,sec:kernel,sec:remedy}, and finally robustness to the training setup.

\paragraph{Protocol.}
We simulate \cref{eq:model} with every source of finite-sample noise included. Each step draws a minibatch of problems, samples $K$ answers per problem, takes the majority vote, and applies the SGD step of \cref{sec:theory}.
Unless stated otherwise we use REINFORCE with $K{=}3$ and train until every problem has left the unstable window.
Initial preferences are exactly balanced, the shared parameter starts at $v{=}0$, and the initial majority is correct for $80\%$ of problems, so collapse can only be driven by sampling noise.
We call a run collapsed when the order parameter $\Phi=|\langle\mathrm{sign}\,z_i\rangle|$, which is $0$ for balanced answers and $1$ when all problems give the same answer, exceeds $0.5$ at the end of training. $N_c$ is the midpoint of a logistic fit of $P(\text{collapse})$ against $\ln N$.
All settings are listed in \cref{app:details}, and code and raw results are available at \url{https://github.com/pd2714/RLVR_instability}.

\begin{figure*}[t]
\centering
\includegraphics[width=\textwidth]{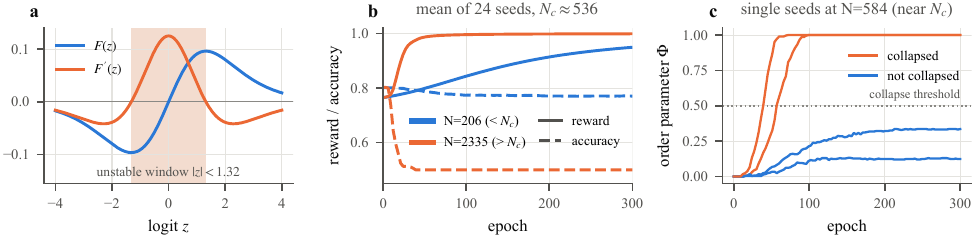}
\caption{\textbf{Mechanism.} (a)~Single-problem drift $F$ and its slope $F'$. The shaded window $|z|<z^\star$ is where a shared push is amplified. (b)~Majority-vote reward (solid) and accuracy against the hidden labels (dashed) at $\lambda{=}0.1$, $a{=}0.5$, averaged over 24 seeds. The two runs differ only in the number of training problems, which lies below (blue) or above (orange) the fitted $N_c\approx536$. Below $N_c$, local sharpening wins: the reward rises and accuracy stays near its initial level. Above $N_c$, the reward rises \emph{faster} and accuracy falls to chance, because every problem converges to the same answer. (c)~Order parameter $\Phi$ of four individual seeds with identical settings at $N=584$, the simulated size closest to $N_c$. Orange runs end above the collapse threshold $\Phi=0.5$ (dotted) and blue runs below it. Sampling noise decides the outcome within the first ${\sim}100$ epochs, while problems are still inside the unstable window, and $\Phi$ then freezes. A run can also collapse partially (upper blue curve).}
\label{fig:mechanism}
\end{figure*}

\subsection{A sharp transition at a single critical amplification (P1--P4)}
\begin{figure*}[t]
\centering
\includegraphics[width=\textwidth]{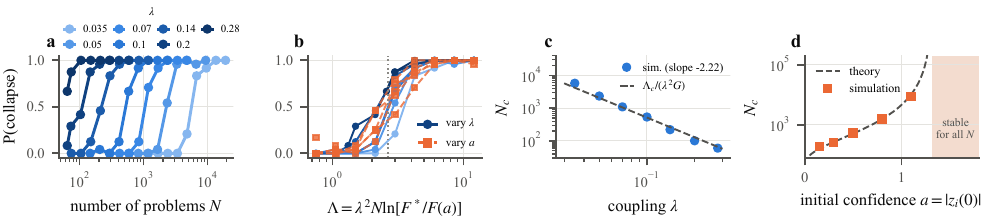}
\caption{\textbf{Critical number of problems.} (a)~Probability of collapse against $N$ for seven coupling strengths ($a{=}0.5$). (b)~The same data, together with a sweep over initial confidence $a$ (orange, $\lambda{=}0.1$), plotted against the predicted amplification $\Lambda=\lambda^2N\ln[F^\star/F(a)]$. Curves spanning a $64\times$ range in $\lambda^{-2}$ and critical sizes from $60$ to $8740$ collapse onto a single transition at $\Lambda_c\approx2.6$ (dotted). (c)~$N_c$ against $\lambda$. The fitted slope is $-2.22$, compared with the predicted $-2$ plus a logarithmic correction. (d)~$N_c$ against initial confidence. The dashed line is the theory with the single constant $\Lambda_c$. Beyond $z^\star$ (shaded) the theory predicts no collapse at any $N$.}
\label{fig:scaling}
\end{figure*}

We begin with the basic phenomenon (P1): whether a sharp transition in $N$ appears without being built into the model.
Above a threshold number of problems, the self-assessed reward goes to one while accuracy falls to chance, because every problem converges to the same answer (\cref{fig:mechanism}b).
Below it, the same algorithm with the same hyperparameters only sharpens each problem.
Near $N_c$, runs with identical settings differ only in their sampling noise, and some collapse while others do not (\cref{fig:mechanism}c).
The outcome is decided early, while problems are still inside the unstable window, and does not change afterwards.
The probability of collapse rises from $0$ to $1$ within a factor of a few in $N$ (\cref{fig:scaling}a).

Having established the transition, we next ask where it occurs (P2). \Cref{eq:Nc} predicts that the coupling $\lambda$ and the initial confidence $a$ enter only through the amplification $\Lambda$.
Varying the coupling moves $N_c$ by two orders of magnitude, with $N_c\propto\lambda^{-2.2}$ (\cref{fig:scaling}c). The small excess over $-2$ is the predicted $\ln(1/\delta)$ correction, since the noise seed scales with $\lambda$ (\cref{app:seed}).
Varying the initial confidence moves $N_c$ as $1/\ln[F^\star/F(a)]$ (\cref{fig:scaling}d).
Plotted against $\Lambda$, all sweeps collapse onto a single transition at $\Lambda_c\approx2.6$ (\cref{fig:scaling}b), and this one constant then predicts $N_c$ for every other configuration below.
Fitted values are in \cref{app:fits}, and \cref{app:mfcheck} checks the form of \cref{eq:Nc} exactly in the noise-free limit.

The last two predictions also hold.
Batch size changes $N_c$ only logarithmically, and not at all under the linear scaling rule $\eta\propto B$ (P3, \cref{app:batch}).
Confident models ($a>z^\star$) never collapse, even far beyond the threshold for uncertain models and under much longer training: not a single problem flips its answer (P4, \cref{app:stable}).

\subsection{Heterogeneity, feature coupling and a remedy}
\label{sec:hetero}
The sweeps above use a homogeneous population with no coupling besides the shared direction. We now test heterogeneous confidence (\cref{sec:general}), prediction-aligned feature coupling (\cref{sec:kernel}) and the marginal-entropy remedy (\cref{sec:remedy}). Details are in \cref{app:hetero}.

\begin{figure*}[t]
\centering
\includegraphics[width=\textwidth]{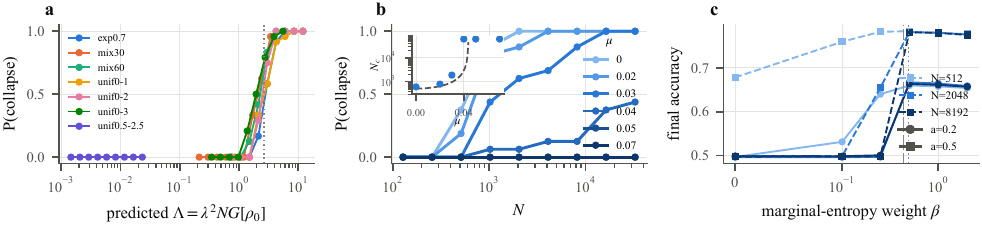}
\caption{\textbf{Generalizations.} (a)~Heterogeneous initial confidence. $P(\text{collapse})$ against the amplification $\lambda^2NG[\rho_0]$ predicted from the initial distribution alone; the dotted line is $\Lambda_c$ from \cref{fig:scaling}. (b)~Prediction-aligned coupling $\mu$ delays and then eliminates collapse. Inset: $N_c$ against $\mu$ with the prediction $1/(\lambda^2G/\Lambda_c-\mu^2)$. Points at the top edge did not collapse. (c)~Final accuracy with the marginal-entropy bonus. Vertical lines mark the predicted $\beta_c(a)$.}
\label{fig:hetero}
\end{figure*}

Real models are uncertain on some problems and confident on others.
For seven distributions of initial confidence, the amplification $\lambda^2NG[\rho_0]$, computed from the initial distribution alone, locates the transition in every case, with the same $\Lambda_c$ (\cref{fig:hetero}a).
What matters is the sign of the instability index $\langle F'\rangle$, not the mean confidence. A population that is mostly confident still collapses, only at a much larger $N$, as long as some problems start inside the window. A population with $\langle F'\rangle<0$ from the start never collapses.

Heterogeneity changes $G$ but leaves the coupling between problems unchanged. We next add the protective coupling predicted by the kernel theory: a coupling aligned with each problem's initial preference, so that sharpening one problem also sharpens the others, as happens when a network generalizes.
The kernel theory (\cref{eq:kernel}) predicts that $N_c$ grows with the strength $\mu$ of this coupling and diverges at a finite $\mu_c$.
Simulations agree (\cref{fig:hetero}b): $N_c$ grows with $\mu$, and beyond $\mu_c$ no run collapses at any $N$ we tested.
When sharpening is shared strongly enough across problems, adding data no longer leads to collapse.

Finally, the marginal-entropy bonus of \cref{sec:remedy} prevents collapse above the predicted threshold $\beta_c\approx\tfrac12$ (\cref{eq:beta}, \cref{fig:hetero}c).
Above it, no run collapses at any $N$ we tested, and because the bonus does not slow per-problem sharpening, accuracy stays at its sharpening-only value.

\subsection{Robustness to the training setup}
All runs above use REINFORCE with $K=3$, binary answers and plain SGD. Varying each of these choices leaves the law intact (\cref{sec:robust}).
GRPO collapses at the same $N_c$ as REINFORCE, as predicted for $K=3$ (\cref{sec:single}). Larger groups raise $N_c$ in the order predicted by $G_K$, many-answer policies collapse at a similar $\Lambda_c$, and Adam keeps $N_c\propto\lambda^{-2}$ with a larger constant (\cref{fig:robust}).

\begin{figure*}[t]
\centering
\includegraphics[width=\textwidth]{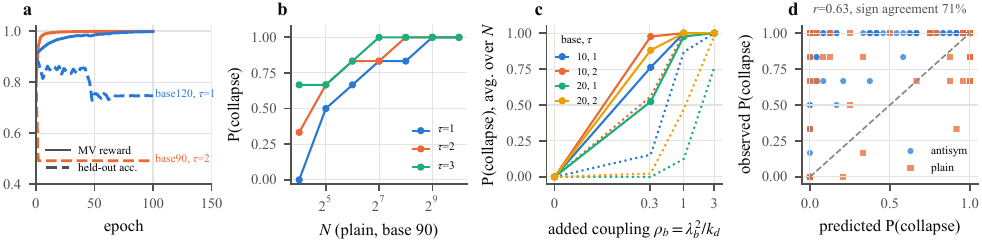}
\caption{\textbf{Label-free RL on a small Transformer.}
(a)~Majority-vote reward (solid) and held-out accuracy (dashed) of one run at $N{=}512$, for base 90 at $\tau{=}2$ (orange) and base 120 at $\tau{=}1$ (blue).
(b)~Plain Transformer, base 90: probability of collapse against $N$ at three temperatures, as the observed fraction of 6 seeds.
(c)~Antisymmetric Transformer: probability of collapse, averaged over $N$, against the added shared coupling $\rho_b$, for each base and $\tau$. Solid: observed. Dotted: kernel-calibrated surrogate in the same color, from 24 simulated seeds per point. $\rho_b{=}0$ is drawn at the left edge.
(d)~Observed against surrogate-predicted probability of collapse. Each point is one condition at one training-set size $N$, summarizing 6 RL runs, so observed values are fractions of 6 and many points overlap. Dashed: perfect calibration; points above it collapse more often than predicted. Title: Pearson correlation $r$ and the fraction of points in which prediction and observation lie on the same side of $\tfrac12$.}
\label{fig:nn}
\end{figure*}

\input{sec_nn}

%% file: sec_nn.tex
\section{A Small Transformer Trained Without Labels}
\label{sec:nn}

\Cref{sec:sims,sec:robust} test the theory on the toy model, where the coupling is known exactly. A real network has many shared directions, and its coupling must be measured.
We therefore ask whether quantities measured on an initial \emph{network}, before any RL step, predict whether label-free RL collapses.
We describe the task and models, form predictions from the initial network, and then compare them with the outcomes of RL.

\subsection{Setup}
\label{sec:nnsetup}

\paragraph{Task and architectures.}
Inputs are six digits, and the answer is $+$ if $x_1+x_2+x_3>x_4+x_5+x_6$ and $-$ otherwise.
The policy is a small Transformer that reads out an answer logit $z(x)$ from a classification (CLS) token (architecture in \cref{app:details}). Answers are sampled at temperature $\tau$, so the logit of \cref{sec:single} is $z/\tau$.
Raising $\tau$ makes the same network less confident, which moves problems into the unstable window and increases $G$, without changing the kernel ratios (every kernel entry scales as $1/\tau^2$). We therefore use $\tau$ to vary the entropy of the base model.
To mimic a pretrained base model with partial competence, we first train briefly with labels. ``Base $n$'' denotes the model after $n$ supervised steps.

We describe the coupling of each network by the kernel ratios of \cref{sec:kernel}: $k_u/k_d'$ measures the shared coupling, which drives collapse, and $k_s/k_d'$ the prediction-aligned coupling, which protects against it. We measure them on the initial network, from per-example gradients on a sample of unlabeled problems.
We study two architectures chosen to sit at opposite ends of \cref{eq:kernel}: one whose shared coupling far exceeds its protective coupling, and one with essentially no shared coupling but strong protection. Together they test the kernel prediction at both extremes, and the second provides a clean baseline into which a controlled shared coupling can be added back.
Let $h(x)$ denote the network's readout, the logit difference computed from the CLS token. The \emph{plain} model uses it directly, $z(x)=h(x)$. Its readout features share a large input-independent component, so its shared coupling is an order of magnitude larger than its protective one.
The \emph{antisymmetric} model uses the same network and parameters but constrains its logit to be antisymmetric under exchanging the two halves of the input, like an odd function: $z(\bar x)=-z(x)$, where $\bar x=(x_4,x_5,x_6,x_1,x_2,x_3)$. We impose this by running the network on both inputs and taking the difference, $z(x)=h(x)-h(\bar x)$. The constraint matches the task, since exchanging the halves always flips the correct answer. It also removes the shared direction: a parameter change that shifts $h$ by the same amount on every input, which in the plain model pushes all problems toward the same answer, cancels in the difference. This leaves essentially no shared coupling while keeping a strong protective one.
For each architecture we use two base models (\cref{tab:conditions}).

In the antisymmetric model we reintroduce a controlled shared coupling through a scaled output bias, $z\mathrel{+}=\lambda_bb$. This adds $\lambda_b^2$ to $k_u$ and leaves $k_s$ and $k_d'$ essentially unchanged.
We measure the dose by $\rho_b=\lambda_b^2/k_d$, relative to the self-kernel $k_d$ of the bias-free model.

\paragraph{RL runs and parameter-free predictions.}
We run label-free majority-vote RL (REINFORCE, $K{=}3$, plain SGD) on the first $N$ problems of an unlabeled pool, with the step scaled so that each visit moves a problem's own logit by the same amount as in the toy model.
We call each combination of architecture, base model, temperature and dose a \emph{condition}, and run each condition at seven training-set sizes with six seeds (\cref{app:nndetails}).

From the initial network alone, we then predict whether training collapses.
Our main predictor is a \emph{kernel-calibrated surrogate}: the stochastic toy model of \cref{sec:sims} with $\lambda^2=k_u/k_d'$, $\mu^2=k_s/k_d'$, the network's own initial logits, and the same training settings (\cref{app:kernel}).
As simpler baselines we use the bare threshold $\Lambda(N)>\Lambda_c$ of \cref{eq:kernel}, with $\Lambda_c$ from the toy model, and a deterministic mean-field version of the kernel dynamics (\cref{app:kernel}).
None of these predictors is fitted to the network's RL outcomes.

\subsection{Results}
\label{sec:nnresults}

\paragraph{Collapse and what controls it.}
Label-free RL reproduces the failure seen in language models (\cref{fig:nn}a). In collapsing runs the majority-vote reward goes to one while training \emph{and held-out} accuracy fall to chance, because the network gives the same answer to every input.
Not every run fails this way: for the most confident plain model the reward also rises, but held-out accuracy stays far above chance.

The two architectures let us test whether this collapse requires a shared direction.
The antisymmetric network, which has none, \emph{never} collapses, as the surrogate predicts (\cref{fig:nn}c). This holds even for highly uncertain initializations, which would collapse at small $N$ given any appreciable shared coupling.
Adding a shared coupling back through the bias restores collapse in a dose-dependent way, and once it is comparable to the self-kernel ($\rho_b\ge1$) virtually every run collapses (\cref{fig:nn}c).
This is the network analogue of the $\lambda$-sweep of \cref{sec:sims}, and it directly tests the causal claim: the instability requires a direction that moves all answers together.

The plain network, which has a strong intrinsic shared direction, shows the two dependences reported for language models.
Its probability of collapse grows with $N$ (\cref{fig:nn}b), the small-set-stable, large-set-unstable pattern of \citet{he2026urlvr}, here at smaller $N$ because its shared coupling is much stronger than in our toy-model runs.
Raising the temperature, which increases the initial entropy, increases collapse monotonically for both base models (\cref{fig:nn}b, \cref{tab:conditions}).

\paragraph{Accuracy of the parameter-free predictions.}
We score each predictor on every condition and training-set size, counting agreement when the predicted and observed probabilities of collapse lie on the same side of $\tfrac12$.
The surrogate agrees in $71\%$ of cases (\cref{fig:nn}d), and more often far from threshold. The bare threshold $\Lambda(N)>\Lambda_c$ ($58\%$) and the deterministic kernel mean-field ($63\%$) do worse (\cref{app:nndetails}).
The surrogate's value lies in ranking rather than calibration. It separates collapsing from stable runs well (area under the ROC curve $0.87$) and has a lower Brier score than either fixed-answer predictor, but on average it predicts collapse far less often than it occurs (\cref{app:nndetails}).
The errors are almost all in one direction: the real network collapses more often than its initial kernel predicts.
Most misses are antisymmetric networks with weak added coupling, which sit close to threshold, and confident plain networks at $\tau=1$, whose initial logits place almost no problems inside the unstable window. The plain base-120 network at $\tau=1$ has initial $G=0.000$, yet $17$ of its $42$ runs collapse (\cref{tab:conditions}).
Predictions from the initial kernel are therefore not sufficient to establish a reliable stability boundary in a real network.

One important source of these errors is that the kernel changes during training, which breaks the fixed-kernel assumption of \cref{sec:kernel}. In some networks feature learning quickly erodes the protective coupling relative to the shared one, and in others the self-kernel grows by orders of magnitude (\cref{app:kerneldrift}). Kernel drift is small in the plain networks, however, and does not account for their misses.

\paragraph{Marginal-entropy remedy.}
The bonus of \cref{sec:remedy} also prevents collapse in the network, but it does not rescue accuracy (\cref{fig:nnremedy}, \cref{app:nnremedy}).
The bonus penalizes only an imbalance of the pooled answers, so it damps only the uniform mode. The kernel has other collective modes that move many problems at once while keeping the answers balanced, and that are unrelated to the current predictions. \Cref{prop:instab} applies to each of them, so once the uniform mode is damped, the strongest of them takes over. The network still ends confident and balanced, but its answers follow a pattern set by its own features rather than by the task, and accuracy falls to chance.
Removing the imbalance of the answer marginal is therefore not the same as preventing collective collapse: the bonus stabilizes the uniform mode, but a task-unrelated collective mode can still take over.

%% file: sec_discussion.tex
\section{Discussion}
\label{sec:discussion}

We close by summarizing the mechanism, drawing practical implications and stating the limitations of our analysis.

\paragraph{Why adding data destabilizes self-rewarded RL.}
Our account of the dataset-size threshold of \citet{he2026urlvr} is a race.
Each problem-specific degree of freedom receives a $1/N$ share of the learning signal, while shared directions receive all of it.
As long as many problems are undecided, the shared directions are dynamically unstable, and the longer it takes to sharpen the individual problems (larger $N$), the more a shared bias is amplified.
The resulting threshold depends on the coupling ($\propto\lambda^{-2}$) and on the initial confidence profile through a single number $G$, and only logarithmically on the batch size and the noise.
That higher-entropy base models collapse faster \citep{he2026urlvr} follows directly, because they have more problems in the unstable window and hence larger $G$.
Our results also make precise the observation of \citet{he2026urlvr} that entropy is a \emph{consequence} of sharpening rather than its cause. In the theory, what matters is the sign of the population-averaged slope $\langle F'\rangle$, not the entropy itself.

\paragraph{Practical implications.}
Beyond explaining the threshold, the theory suggests concrete diagnostics and levers.
(1)~\emph{Measure before training.} The index $\langle F'(z_i)\rangle$ over the training set and the kernel ratios $k_u/k_d'$ and $k_s/k_d'$ can be estimated on the initial model from samples and per-example gradients. In the solvable model they predict $N_c$ (\cref{sec:sims}). In a feature-learning network the initial kernel ranks the risk but tends to underestimate it, partly because the kernel changes during training (\cref{tab:drift}), so it is a screening tool rather than a stability guarantee.
(2)~\emph{Reduce shared coupling.} Anything that removes answer-independent shared directions lowers $\lambda$. Our antisymmetric readout is an extreme example (\cref{fig:nn}c), and answer-format normalization or freezing output biases are milder ones.
(3)~\emph{Stabilize the uniform mode directly.} A marginal-entropy bonus above $\beta_c\approx\tfrac12$ (\cref{eq:beta}) stabilizes the uniform mode without slowing per-problem sharpening (\cref{fig:hetero}c, \cref{fig:nnremedy}). It biases the answer marginals toward balance, which is inappropriate when the true answer distribution is very skewed. In networks it must be complemented by damping the other collective kernel modes that are not aligned with current predictions (\cref{app:nnremedy}).
(4)~\emph{Curriculum by confidence.} Training first on problems on which the model is already confident ($|z|>z^\star$) or in small chunks keeps $\lambda^2NG$ below threshold. This follows from P4 and \cref{eq:Lambda}, but we have not tested it.
Larger groups $K$ narrow the unstable window but only shift $N_c$ moderately (\cref{tab:window}, \cref{fig:robust}a).

\paragraph{Limitations.}
These implications rest on simplifications that we now make explicit.
We model answer-level policies and linearize the coupling between problems. Real language models have many structured shared directions that change during training.
\Cref{prop:instab} describes the instantaneous growth of a shared perturbation along the sharpening trajectory, not the stability of a fixed point, and it is not by itself sufficient for collapse. The integrated criterion $\Lambda>\Lambda_c$ and the stability of confident populations (P4) are derived for the homogeneous fixed-coupling model.
In the Transformer they do not transfer quantitatively: a network with initial $G=0.000$ still collapses in 17 of 42 runs (\cref{tab:conditions}), and predictions from the initial kernel underestimate the risk and agree with outcomes no more often than always predicting collapse (\cref{app:nndetails}). The kernel drifts (\cref{tab:drift}): in some networks the protective coupling erodes, and the self-kernel can grow by two orders of magnitude. The network also has collective modes besides the uniform one, and damping only the uniform mode does not prevent convergence to an arbitrary balanced rule.
$\Lambda_c$ is calibrated on the toy model and depends logarithmically on the noise seed.
Because of compute limits we did not test pretrained language models. Measuring $\langle F'\rangle$ and the kernel ratios on a small pretrained language model, and testing whether they predict the dataset-size threshold of \citet{he2026urlvr}, is the most important next step.

\paragraph{Conclusion.}
Unsupervised RLVR is a many-body system: the problems interact through shared parameters.
Majority-vote self-reinforcement turns the balanced state, in which different problems keep different answers, into a paramagnet that amplifies any shared bias while too many problems are undecided at once.
This single mechanism explains why collapse appears only above a critical number of problems, predicts how that number scales with coupling, confidence and batch size, identifies a protective counterpart in generalizing (prediction-aligned) coupling, and shows how to stabilize the uniform collective mode, together with why that alone does not preserve accuracy in a network.

%% file: sec_appendix.tex
\input{sec_appendix_theory}

\section{Further Results for the Solvable Model}
\label{app:simdetails}
This section collects the numbers behind \cref{sec:sims}.

\subsection{Fitted critical sizes (P1, P2)}
\label{app:fits}
The default settings are REINFORCE, $K{=}3$, $B{=}32$, SGD with step $\eta{=}4$ and 300 epochs, so each visit changes $u_i$ by $(\eta/B)\,\hat g_i=0.125\,\hat g_i$. Each configuration is run with 16--24 seeds.
In collapsing runs the self-assessed reward goes to $0.999$ while accuracy falls from $0.80$ to $0.50$ (\cref{fig:mechanism}b). Near $N_c$ the outcome is fixed within the first ${\sim}100$ epochs (\cref{fig:mechanism}c), and the probability of collapse rises from $0$ to $1$ within a factor of about $3$--$4$ in $N$ (\cref{fig:scaling}a).
Varying only the coupling, $\lambda\in[0.035,0.28]$, moves the critical size from $N_c=5750$ to $N_c=60$, and a power-law fit gives $N_c\propto\lambda^{-2.22}$ (\cref{fig:scaling}c).
Varying only the initial confidence, $a\in[0.15,1.1]$, moves $N_c$ from $189$ to $8740$ (\cref{fig:scaling}d).
The fitted critical amplification is $\Lambda_c=2.64$, and the twelve sweeps (seven couplings and five confidences) give values in $[2.1,3.6]$ (\cref{fig:scaling}b).

\begin{figure}[t]
\centering
\includegraphics[width=\textwidth]{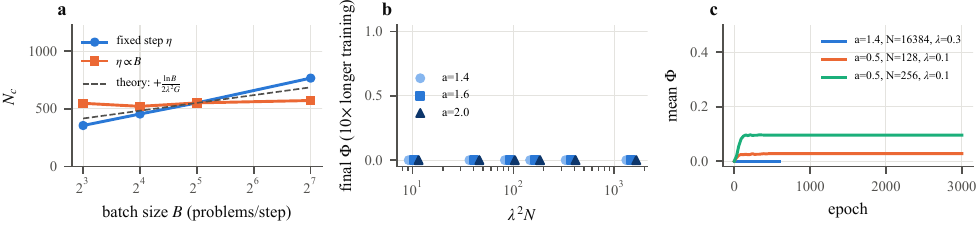}
\caption{\textbf{Batch size and long training.} (a)~$N_c$ against batch size, with a fixed step (blue) and with linear step scaling (orange). The dashed line is the logarithmic prediction anchored at $B{=}32$. (b)~Final order parameter after $10\times$ longer training for confident initial models ($a>z^\star$) at values of $\lambda^2N$ up to $1474$. Uncertain models collapse for $\lambda^2N\gtrsim5$. (c)~Order-parameter trajectories: confident models stay balanced, and sub-critical uncertain models remain balanced after sharpening.}
\label{fig:batch}
\end{figure}

\subsection{Batch size enters logarithmically (P3)}
\label{app:batch}
The remaining input to \cref{eq:Nc} is the noise seed $\delta$, which the batch size controls.
With a fixed step size, a $16\times$ change in batch size ($B=8\to128$) changes $N_c$ by only $2.2\times$ ($355\to765$, \cref{fig:batch}a), roughly linearly in $\ln B$ as predicted. The crude seed estimate in \cref{app:derivations} gives a slope of $0.5/(\lambda^2G)$ per unit $\ln B$; we measure about $1.5\times$ that.
For $B=512$ every tested size ($N\ge2B$) already collapses, consistent with $N_c<1024$.
Under the linear scaling rule $\eta\propto B$, which keeps the noise per sample fixed, $N_c$ is flat ($521$--$572$ for $B=8$--$128$, orange in \cref{fig:batch}a), again as predicted by \cref{app:seed}.
At $B=512$ with linear scaling the step on the shared parameter ($\eta\lambda^2=0.64$) is too large for the small-step approximation, and all runs collapse.
For comparison, a $16\times$ change in $\lambda^{-2}$ moves $N_c$ by $18$--$26\times$ (\cref{fig:scaling}c).

\subsection{Confident models are stable, not merely slow (P4)}
\label{app:stable}
The last prediction concerns models that start outside the unstable window, which the theory says are stable rather than merely slow to collapse.
To distinguish stability from slow dynamics, we train confident models, $a\in\{1.4,1.6,2.0\}>z^\star$, at couplings and sizes far beyond the uncertain-model threshold: up to $\lambda^2N=1474$, where an $a=0.5$ population would reach $\Lambda\approx760$.
For $N=1024$ and $4096$ we train $10\times$ longer (3000 epochs), and for $N=16384$ $2\times$ longer.
Across all 18 configurations (16 seeds each) \emph{not a single problem flips} its answer. The shared shift stays at $|\lambda v|\le0.013$, and accuracy stays at its initial $0.80$ (\cref{fig:batch}b,c).
Sub-critical uncertain models ($a=0.5$, $N\le256$, $\lambda^2N\le2.6$) also remain uncollapsed under $10\times$ longer training ($\Phi\le0.1$ on average, \cref{fig:batch}c).
Once the population has left the unstable window, the shared mode is damped, so training longer does not lead to collapse.

\subsection{Heterogeneity, feature coupling and remedy}
\label{app:hetero}
\paragraph{Heterogeneous confidence.}
We draw $|z_i(0)|$ from seven distributions: uniform on $[0,1]$, $[0,2]$, $[0,3]$ and $[0.5,2.5]$; exponential with mean $0.7$; and two-point mixtures with $30\%$ or $60\%$ of problems at $0.3$ and the rest at $2.0$.
For each we compute $G[\rho_0]$ from the initial distribution alone, using its definition in \cref{sec:general}.
The measured $N_c$ spans $455$--$14{,}700$, and $\Lambda_c\in[2.0,2.8]$ throughout (\cref{fig:hetero}a).
In the 30\% mixture most problems are confident, yet the population still collapses, at $N_c\approx14{,}700$, consistent with its small but positive $G=0.014$.
For the uniform $[0.5,2.5]$ population, $\langle F'\rangle<0$ from the start, so $G=0$, and no run collapses even at $N=24{,}000$.

\paragraph{Feature coupling.}
We add a coupling $z_i\mathrel{+}=\mu s_iw$ aligned with each problem's initial preference $s_i$. The kernel theory (\cref{eq:kernel}) predicts $N_c=1/(\lambda^2G/\Lambda_c-\mu^2)$, which diverges at $\mu_c=\sqrt{\lambda^2G/\Lambda_c}$. With $\lambda=0.1$, $G(0.5)=0.514$ and $\Lambda_c=2.64$ this gives $\mu_c=0.044$.
The transition broadens near the divergence (\cref{fig:hetero}b). For $\mu=0.04$ only $44\%$ of runs collapse even at $N=32768$, and for $\mu\ge0.05$ no run collapses at any $N$ up to $32768$.

\paragraph{Marginal-entropy remedy.}
\Cref{eq:beta} gives $\beta_c(0.2)=0.49$ and $\beta_c(0.5)=0.44$.
At $\beta=0.5$ and above, no run collapses for any $N\le8192$ or either $a$, and accuracy stays at its sharpening-only value ($0.78$ for $a=0.5$). At $\beta=0.25$, runs survive at $N=512$ but collapse at $N=8192$ (\cref{fig:hetero}c).
At $\beta\ge\beta_c$ the final confidence $\langle|z|\rangle$ is the same as in sub-critical runs without the bonus.

\input{sec_appendix_robust}

\section{Transformer: Detailed Results}
\label{app:nndetails}
This section collects the numbers behind \cref{sec:nn}.

\subsection{Models and conditions}
Ties $x_1+x_2+x_3=x_4+x_5+x_6$ are excluded, so the labels are balanced.
The plain model has $k_u/k_d'=0.40$--$0.70$ and $k_s/k_d'\approx0.05$. We use base 90 and base 120, with $82\%$ and $89\%$ held-out accuracy.
The antisymmetric model has $k_u/k_d'<0.002$ and $k_s/k_d'=0.16$--$0.45$. We use base 10 and base 20, with $84\%$ and $92\%$ held-out accuracy.
The added coupling takes the values $\rho_b\in\{0,0.3,1,3\}$; for example, $\rho_b=1$ gives $k_u/k_d'\approx1.2$--$1.5$.
The plain models are run at $\tau\in\{1,2,3\}$, and the antisymmetric ones at $\tau\in\{1,2\}$ and every $\rho_b$, which gives $6+16=22$ conditions (\cref{tab:conditions}).
Each condition is run at 7 values of $N$ (16--1024 for the plain model, 32--2048 for the antisymmetric one) with 6 seeds, for 924 RL runs in total.
The bare threshold uses $\Lambda_c=2.64$ from the toy model (\cref{fig:scaling}b).

\subsection{Collapse counts}
In collapsing runs the majority-vote reward reaches $\ge0.999$ while held-out accuracy falls from $0.82$ to $0.50$ (\cref{fig:nn}a). The run that does not collapse in \cref{fig:nn}a is base 120 at $\tau=1$.
Without added coupling, the antisymmetric network collapses in $0$ of $168$ runs, across both bases, both temperatures and every $N$ up to $2048$. This includes base 10 at $\tau=2$, with $G=1.12$. With $\rho_b\ge1$, $335$ of $336$ runs collapse.
For the plain base 90 at $\tau=1$, collapse rises from $0/6$ at $N=16$ to $6/6$ for $N\ge512$. Its shared coupling $k_u/k_d'=0.70$ is far larger than $\lambda^2=0.01$ in the toy-model runs of \cref{fig:mechanism}.
Across temperatures $\tau=1,2,3$, base 90 collapses in $29$, $34$ and $37$ of 42 runs, and base 120 in $17$, $35$ and $38$.
At very small $N$ the finite-sample imbalance of the initial answers (up to $\Phi_0=0.38$ for 16--64 problems) acts as a large seed $\delta$, which produces some collapse even at small $N$.

\subsection{Scoring the predictions}
We score each \emph{cell}, a condition at one training-set size. There are $22\times7=154$ cells of 6 runs each. A cell's observed outcome is the fraction of its runs that collapse, and a predictor agrees with it when both lie on the same side of $\tfrac12$.
The kernel-calibrated surrogate agrees in $71\%$ of cells (Pearson $r=0.63$), the deterministic kernel mean-field in $63\%$, and the bare threshold $\Lambda(N)>\Lambda_c$ in $58\%$. Far from threshold ($\rho_b=0$ or $\rho_b\ge1$) the surrogate agrees in $86\%$ of cells, against $67\%$ for always predicting collapse.
In $43$ cells the network collapses although the surrogate predicts stability, and in only $1$ cell the reverse.
Thirty of the 43 are antisymmetric cells in which the added shared coupling is at most about three times the protective one ($k_u\lesssim3k_s$: $\rho_b=0.3$, and base 20 at $\rho_b=1$).
Eight are the confident plain networks at $\tau=1$, where $G\le0.025$ and the surrogate predicts little collapse. Base 90 nevertheless collapses in most runs for $N\ge64$, and base 120 at some sizes but not at others.
The remaining five are base 120 at $\tau=2$ (four cells) and $\tau=3$ (one cell), where the network keeps collapsing at large $N$ but the surrogate predicts stability.
For base 90 at $\tau=2$ and $3$ the surrogate matches the observed collapse at large $N$. At small $N$ it overpredicts collapse, because the measured coupling already places $\Lambda$ near or above $\Lambda_c$ at $N=16$; the one cell predicted to collapse that does not is base 90 at $\tau=2$, $N=16$.

\paragraph{Comparison with trivial baselines.}
Sign agreement has to be read against the base rate. Of the 924 runs, $71\%$ collapse, and $73\%$ of cells have a majority of collapsed runs, so a predictor that always answers ``collapse'' agrees in $73\%$ of cells, slightly more than the surrogate.
The cells are also noisy and correlated. With 6 runs per cell, an observed fraction has a binomial standard error of up to $0.2$, so cells near threshold are classified with large uncertainty, and the 7 cells of a condition share the same network. Resampling the 22 conditions (bootstrap, 5000 draws) gives $95\%$ intervals of $[56\%,86\%]$ for the surrogate's agreement and $[56\%,88\%]$ for always-collapse, and $[-25,+25]$ percentage points for their difference. Sign agreement therefore does not distinguish the surrogate from the trivial predictor.
Probabilistic scores on the 924 individual runs are more informative. The Brier score of the surrogate is $0.235$, better than always-collapse ($0.289$), always-stable ($0.711$), the deterministic kernel mean-field ($0.380$) and the bare threshold ($0.419$). It is worse than a constant prediction equal to the observed collapse rate ($0.205$), although that reference is fitted to the outcomes and therefore optimistic.
The surrogate ranks runs well (area under the ROC curve $0.87$) but is poorly calibrated: its mean predicted probability of collapse is $0.46$, against an observed $0.71$.
Most of its discriminating power comes from separating the antisymmetric networks without added coupling, which it correctly predicts never collapse (all 28 cells), from those with added coupling. Within the plain networks it agrees in $67\%$ of cells, below always-collapse ($81\%$).
The plain base-120 network at $\tau=1$ makes the limitation concrete. Its initial amplification is $G=0.000$ to three decimals, so the fixed-kernel theory predicts essentially no growth of the shared mode, yet $17$ of its $42$ runs collapse (\cref{tab:conditions}).
In short, the initial kernel identifies which architectures are protected and orders conditions by risk, but it does not locate a reliable quantitative stability boundary for a given network.

\subsection{Kernel drift}
\label{app:kerneldrift}
One important source of the one-sided errors is that the kernel changes during training (\cref{tab:drift}). The cases in \cref{tab:drift} do not, however, follow a single pattern.
In the antisymmetric base-20 network, the protective coupling $k_s/k_d'$ drops from $0.42$ to $0.07$ within the first epoch, while the shared coupling $k_u/k_d'$ only halves, so the saturation level $Gk_u/k_s$ of $\Lambda$ roughly triples. Here feature learning erodes the protective coupling relative to the shared one, as the static theory would need to anticipate.
In the antisymmetric base-10 network the dominant change is different: the self-kernel $k_d$ grows by $28$--$245\times$ in a few epochs, so the effective step leaves the small-step regime altogether, and both ratios move mainly because of this growth ($k_s/k_d'$ even rises at first).
The plain networks drift much less: $k_d$ changes by at most $1.6\times$ over the first 4 epochs and $2.2\times$ over 8, and in base 90 $k_s/k_d'$ increases rather than decreases. Kernel drift therefore does not explain why the confident plain networks at $\tau=1$ collapse more often than predicted. Collective kernel modes other than the uniform one (\cref{app:nnremedy}) and the finite-sample imbalance of the initial answers at small $N$ are plausible contributors there, which we have not separated.
A predictor for strongly feature-learning networks must track the kernel, and in our runs the initial kernel tends to underestimate the risk of collapse.

\subsection{Remedy and the next collective mode}
\label{app:nnremedy}
\begin{figure}[t]
\centering
\includegraphics[width=0.6\textwidth]{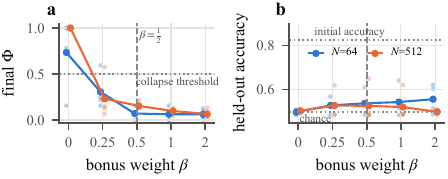}
\caption{\textbf{Marginal-entropy bonus on the plain base-90 Transformer at $\tau=2$.} Lines: mean over 4 seeds; dots: individual seeds. The dashed line marks $\beta=\tfrac12$, above which the theory predicts stability (\cref{eq:beta}). (a)~Final order parameter $\Phi$ on the training problems; runs above the dotted line count as collapsed. The bonus removes collapse for $\beta\ge\tfrac12$. (b)~Held-out accuracy stays near chance at every $\beta$, far below the base model's initial accuracy (dotted, top).}
\label{fig:nnremedy}
\end{figure}

We test the marginal-entropy remedy of \cref{sec:remedy} on the plain base-90 network at $\tau=2$, which collapses at every $N$ without intervention (\cref{fig:nnremedy}).

Without the bonus $7/8$ runs collapse, at $\beta=0.25$ $2/8$ do, and for $\beta\in\{0.5,1,2\}$ none of 24 do, matching the predicted threshold $\beta_c\le\tfrac12$ of \cref{eq:beta}.
The bonus does not, however, rescue accuracy. Remedied runs end with balanced answers ($\Phi\approx0.1$) and high confidence ($\langle|z|\rangle\approx7$), but at chance accuracy, and the learned labelling agrees with no simple input feature better than $58\%$.
The kernel explains this as well. Besides the uniform mode (eigenvalue $1.45\times10^4$), the initial kernel has several collective modes orthogonal to both the uniform direction and the current predictions, with eigenvalues $5.1\times10^3$, $2.5\times10^3$, $2.3\times10^3,\dots$, all far above the self-kernel $k_d=94$.
\Cref{prop:instab} applies to any such mode: when the uniform mode is damped, the next one takes over, and the network collapses onto a balanced but arbitrary rule.
A complete remedy must therefore damp every collective direction not aligned with the current predictions, not only the uniform one. Our analysis identifies these directions as the leading eigenvectors of the kernel restricted to that subspace.

\input{tab_conditions}
\input{tab_nn}

\section{Experimental Details}
\label{app:details}

\paragraph{Solvable model.}
All simulations use the vectorized simulator in \texttt{src/rlvr\_collapse/sim.py}, which runs 16--24 independent seeds in parallel.
Each optimizer step draws $B$ problems without replacement (a fresh permutation each epoch), samples $K$ answers per problem from $\sigma(z_i)$ (or the softmax for $M>2$), breaks majority ties uniformly at random, and applies the chosen advantage.
Each step moves $u_i$ by $\eta\hat g_i/B$ and $v$ by $\eta\lambda\sum_i\hat g_i/B$, where $\hat g_i$ is the per-problem estimate of $F(z_i)$ (\cref{sec:theory}).
\Cref{tab:hparams} lists the settings for every sweep.
$N_c$ is obtained by maximum-likelihood fitting of $P(\text{collapse})=\sigma(k(\ln N-\ln N_c))$ to the binomial counts.
The mean-field and theory computations are in \texttt{theory.py} and \texttt{multiclass.py}.

\begin{table}[h]
\centering\small
\caption{Simulation settings. Unless listed, $K{=}3$, REINFORCE, $B{=}32$, SGD step $\eta{=}4$, 300 epochs, 24 seeds, and $80\%$ of initial majorities correct.}
\label{tab:hparams}
\begin{tabular}{lll}
\toprule
Sweep & Varied & Notes \\
\midrule
coupling & $\lambda\in\{0.035,\dots,0.28\}$, 9 values of $N$ each & $a=0.5$ \\
confidence & $a\in\{0.15,0.3,0.5,0.8,1.1,1.25\}$ & $\lambda=0.1$, $N\le32768$ \\
batch, fixed step & $B\in\{8,16,32,128,512\}$ & epochs $=300B/32$ (equal step count) \\
batch, linear scaling & $B\in\{8,\dots,512\}$ & $\eta=4B/32$, 300 epochs \\
long training & $a\in\{1.4,1.6,2.0\}$, $\lambda\in\{0.1,0.3\}$ & 3000 epochs (600 at $N=16384$); 16 seeds \\
heterogeneous & 7 initial distributions & $\lambda=0.1$ \\
feature coupling & $\mu\in\{0,\dots,0.07\}$, $N\in[128,32768]$ & $\lambda=0.1$, 16 seeds \\
remedy & $\beta\in\{0,0.1,0.25,0.5,1,2\}$, $a\in\{0.2,0.5\}$ & $\lambda=0.1$, 16 seeds \\
GRPO / $K$ & $K\in\{3,5,8,16\}$; GRPO, mean baseline, REINFORCE & step set so that $\eta F^\star_K\in[0.38,0.48]$ \\
many answers & $(M,K)\in\{(3,3),(4,5),(6,5)\}$ & softmax logits, vector $v$ \\
Adam & $\lambda\in\{0.2,\dots,1.0\}$, $N\in[64,16384]$ & lr $0.01$, $(\beta_1,\beta_2)=(0.9,0.999)$, 16 seeds \\
\bottomrule
\end{tabular}
\end{table}

\paragraph{Transformer.}
Embedding width 64, 2 pre-LN encoder layers, 4 heads, MLP width 256, no dropout, a learned positional embedding and a CLS token. The readout is a bias-free linear map to the logit difference, plus the optional scaled scalar bias $\lambda_bb$.
Supervised warm-up uses Adam (lr $10^{-3}$, batch 128) on fresh labeled samples.
RL uses the first $N$ problems of a fixed pool of 8192 unlabeled problems, plain SGD with step $0.125B/k_d$ (where $k_d$ is the mean self-kernel at the given temperature), $B=16$, $K=3$, REINFORCE with majority-vote rewards, and 100 epochs.
Accuracy is measured on 2000 held-out problems.
Per-example gradients for the kernel statistics are computed with \texttt{torch.func} on 384 pool problems.
The amplification $G$ is computed from the initial logits of 2048 pool problems.
A run counts as collapsed if $|\langle\mathrm{sign}\,z\rangle|>0.5$ on its training problems at the end of training.

\paragraph{Compute.}
All experiments ran on a single Apple M4 Pro CPU (14 cores).
The full solvable-model study takes under an hour, and the 924 Transformer runs take about 40 minutes.

%% file: sec_appendix_theory.tex
\section{Derivations}
\label{app:derivations}

\subsection{Exact drift for general $K$ and estimators}
\label{app:drift}
Let $k$ be the number of $+$ answers among $K$ samples, $k\sim\mathrm{Bin}(K,p)$.
Given $k$ and the (tie-broken) majority $\hat a$, every sample with answer $c\in\{+,-\}$ receives the same reward $r_c=\mathbf 1[c=\hat a]$ and hence the same advantage $A_c$. The summed score-function gradient is $kA_+(1-p)-(K-k)A_-p$.
Averaging over tie-breaks gives coefficients $c_+(k)=\frac kK\mathbb E[A_+\mid k]$ and $c_-(k)=\frac{K-k}K\mathbb E[A_-\mid k]$, and
\begin{equation*}
  F(z)=\sum_{k=0}^K\binom Kk p^k(1-p)^{K-k}\,\big[c_+(k)(1-p)-c_-(k)p\big].
\end{equation*}
Write $\bar r=\max(k,K-k)/K$ for the mean reward when there is no tie. REINFORCE uses $A_c=r_c$, the mean baseline uses $A_c=r_c-\bar r$, and GRPO uses $A_c=(r_c-\bar r)/\sqrt{\bar r(1-\bar r)}$, with $A_c=0$ when all samples agree.
For $K=3$ the estimators split the drift differently between unanimous groups and $2{:}1$ splits.
For REINFORCE, a unanimous group has $r_j=1$ for all samples, so $\hat g=1-p$ for $k=3$ and $\hat g=-p$ for $k=0$. A $2{:}1$ split rewards only its two majority samples, so $\hat g=\tfrac23(1-p)$ for $k=2$ and $\hat g=-\tfrac23p$ for $k=1$. Splitting the expectation accordingly,
\begin{equation*}
  \mathbb E[\hat g\,\mathbf 1_{\rm unanimous}]=p^3(1-p)-(1-p)^3p=p(1-p)(2p-1)=F(z),\qquad
  \mathbb E[\hat g\,\mathbf 1_{2:1}]=2p^2(1-p)^2-2p^2(1-p)^2=0,
\end{equation*}
so for REINFORCE the unanimous groups carry the entire mean drift.
For the mean baseline and GRPO the advantage vanishes in unanimous groups, and only the $2{:}1$ splits contribute. There $\bar r=2/3$, the majority samples receive $A=\tfrac13$ and the minority sample $A=-\tfrac23$. Because the two scores differ by exactly $(1-p)-(-p)=1$, the mean-baseline estimate is $\hat g=\pm\tfrac29$ independently of $p$, with the sign of the majority. Hence
\begin{equation*}
  F_{\rm mean}(z)=\tfrac29\big[3p^2(1-p)-3p(1-p)^2\big]=\tfrac23\,p(1-p)(2p-1)=\tfrac23F(z).
\end{equation*}
GRPO divides these advantages by the common standard deviation $\sqrt{\bar r(1-\bar r)}=\sqrt2/3$, which gives $F_{\rm GRPO}=\tfrac{3}{\sqrt2}\cdot\tfrac23F=\sqrt2F$. The three drifts are therefore proportional, which proves the claim in \cref{sec:single}, even though they draw their signal from different groups.
The same enumeration over multinomial count vectors gives the $M$-answer drift $g(z)\in\mathbb R^M$, whose components sum to zero.

\subsection{Linear stability and the amplification integral}
\label{app:stability}
Write $\epsilon=\eta t/N$ in \cref{eq:mf}.
Let the problems sit at $\pm a(t)$ with $da/d\epsilon=F(a)$, and let $w=\lambda v$ be a common logit shift.
Then $z_i=\pm a+w$ and $\langle F\rangle=\tfrac12[F(a+w)-F(a-w)]=F'(a)w+O(w^3)$, so
\begin{equation*}
  \frac{dw}{d\epsilon}=\lambda^2N\,F'(a)\,w\quad\Rightarrow\quad
  \ln\frac{w(\epsilon)}{w(0)}=\lambda^2N\int_0^\epsilon F'(a)\,d\epsilon'=\lambda^2N\ln\frac{F(a(\epsilon))}{F(a(0))}.
\end{equation*}
The amplification is largest at the window edge $a=z^\star$, which gives $\Lambda=\lambda^2N\ln[F^\star/F(a_0)]$.
After that point $F'<0$ and the shift is damped.
The per-problem term $\frac{\eta}{N}F'(z_i)$ in the full Jacobian amplifies differences between problems on the same side. This only reparametrizes time and cannot flip a problem.
The minority side flips once $w$ exceeds $a(t)$, which gives the threshold $\Lambda\approx\ln(1/\delta)+O(1)$.

For a heterogeneous population, each problem follows $dz_i/d\epsilon=F(z_i)$ to leading order, and the shift grows at rate $\lambda^2N\langle F'(z(\epsilon))\rangle$. The maximal log-amplification is therefore $\lambda^2N\sup_\epsilon\int_0^\epsilon\langle F'\rangle d\epsilon'$.

\subsection{Size of the noise seed}
\label{app:seed}
With minibatches of $B$ problems and step $\eta$, the shared shift receives zero-mean noise of standard deviation $\eta\lambda^2\sigma/\sqrt B$ per step, where $\sigma^2$ is the variance of the single-problem gradient estimate.
Noise injected within the first growth time $t_{\rm g}=1/(\eta\lambda^2F'(a))$ is amplified almost fully, which gives an effective seed $\delta^2\simeq t_{\rm g}\,\eta^2\lambda^4\sigma^2/B=\eta\lambda^2\sigma^2/(BF'(a))$ and $\ln(1/\delta)=\tfrac12\ln B+\ln(1/\lambda)-\tfrac12\ln\eta+\text{const}$.
Under the linear scaling rule $\eta\propto B$ the dependence on $B$ cancels.

\subsection{Mean-field check of the closed form}
\label{app:mfcheck}
We check \cref{eq:Nc} in the noise-free limit, where it can be tested exactly.
In rescaled time $s=\eta t/N$, the deterministic two-population version of \cref{eq:mf} reads $dz_\pm/ds=F(z_\pm)+\lambda^2N\langle F\rangle$, where $z_+$ and $z_-$ are the logits of the problems that start at $+a$ and $-a$ (without sampling noise, all problems in each group stay identical) and $\langle F\rangle=\tfrac12[F(z_+)+F(z_-)]$. It depends on $\lambda$ and $N$ only through $X=\lambda^2N$, so in mean field $N_c\propto\lambda^{-2}$ holds exactly.
Solving for the critical $X_c(a,\delta)$ by bisection, the exact critical amplification $X_cG(a)$ equals $\ln(1/\delta)-c$, with unit slope in $\ln(1/\delta)$ over $\delta\in[10^{-6},10^{-2}]$ and a weakly $a$-dependent offset $c\in[0.9,2.0]$ for $a\in[0.3,1.25]$ ($c\approx1.1$ at $a=0.5$; \cref{tab:mf}). This is the form asserted by \cref{eq:Nc}.
The constant offset is the $O(1)$ threshold $w\sim a$ required for the minority side to flip.
In the stochastic simulations of \cref{sec:sims} we measure $\Lambda_c\approx2.6$ at $a=0.5$ (\cref{fig:scaling}b). Inverting $\Lambda_c=\ln(1/\delta_{\rm eff})-c$ with $c\approx1.1$ gives an effective noise seed $\delta_{\rm eff}\approx e^{-3.7}\approx0.02$.

\input{tab_meanfield}

\subsection{Many answers}
\label{app:multi}
Let $z_i=a\,e_{s_i}$ with labels $s_i$ balanced over $\{1,\dots,M\}$, and let $w=\lambda v\in\mathbb R^M$.
Then $\dot w=\eta\lambda^2\langle g(z_i+w)\rangle$, whose linearization is $\eta\lambda^2\bar J w$ with $\bar J=\frac1M\sum_s\partial g/\partial z\,|_{ae_s}$.
Averaging over $s$ makes $\bar J$ permutation-invariant, so on the zero-sum subspace (the only identifiable directions under softmax) $\bar J=\kappa(a)I$ with $\kappa=\mathrm{tr}(P\bar JP)/(M-1)$ and $P=I-\frac1M\mathbf1\mathbf1^\top$.
The margin $a$ between the preferred and the other logits obeys $\dot a=\frac\eta N f(a)$ with $f=g_{s}-g_{\neq s}$.
Dividing the two rates gives $\Lambda=\lambda^2N\int_{a_0}^{a^\star}\kappa/f\,da$.
For $M=2$ this reduces exactly to the binary result; we checked this numerically.

\subsection{Kernel form}
\label{app:kernel}
In a linearized network, $\dot z=\frac\eta N\mathbf K F(z)$.
Take $\mathbf K=k_d'\mathbf I+k_u\mathbf1\mathbf1^\top+k_sss^\top$ with balanced signs $s$, so that $\mathbf1\perp s$, and a homogeneous population $z_i=s_ia$.
Then $\dot z_i=\frac\eta N(k_d'+Nk_s)F(a)s_i$, so $\dot a=\frac\eta N(k_d'+Nk_s)F(a)$: sharpening is accelerated by the prediction-aligned coupling.
A common shift $z_i=s_ia+w$ gives $F(z_i)=s_iF(a)+F'(a)w+O(w^2)$, since $F$ is odd and $F'$ even. With $\mathbf K\mathbf1=(k_d'+Nk_u)\mathbf1$ this gives $\dot w=\frac\eta N(k_d'+Nk_u)F'(a)w$.
Dividing by $\dot a$ eliminates time:
\begin{equation*}
  \frac{d\ln w}{da}=\frac{k_d'+Nk_u}{k_d'+Nk_s}\,\frac{F'(a)}{F(a)}.
\end{equation*}
Integrating from $a_0$ to $z^\star$ gives
\begin{equation*}
  \Lambda=G(a_0)\,\frac{Nk_u}{k_d'+Nk_s}+G(a_0)\,\frac{k_d'}{k_d'+Nk_s}.
\end{equation*}
The second term is the self-term. It is at most $G$ and independent of $N$, and it is the growth that remains with the coupling switched off, when each problem follows its own sign-preserving flow and cannot flip. It is therefore absorbed into the $O(1)$ of the threshold, as in the toy model (\cref{app:stability}), and dropping it gives \cref{eq:kernel}. The $k_d'$ in the denominator sets the sharpening rate and multiplies the leading term, so it is kept.
We estimate $k_d$ as the mean diagonal of the empirical kernel, $k_u$ as the mean off-diagonal, and $k_s$ as the mean off-diagonal of $s_is_j\mathbf K_{ij}$, and set $k_d'=k_d-k_u-k_s$.

\paragraph{Kernel mean-field predictor.}
The deterministic baseline of \cref{sec:nn} integrates the projected dynamics directly. Writing $\epsilon=\eta k_d't/N$ and $\langle sF\rangle=\frac1N\sum_js_jF(z_j)$,
\begin{equation*}
  \frac{dz_i}{d\epsilon}=F(z_i)+\frac{Nk_s}{k_d'}\,s_i\langle sF\rangle+\frac{Nk_u}{k_d'}\,\langle F\rangle,
\end{equation*}
integrated from the measured initial logits over the same training horizon as the RL runs.

\paragraph{Kernel-calibrated surrogate.}
The main predictor of \cref{sec:nn} keeps the same projected kernel but replaces the deterministic dynamics by the stochastic simulator of \cref{sec:sims}, in units where the problem-specific self-kernel is $k_d'=1$.
For each condition and training-set size $N$ we
(i)~set $u_i=z_i(0)/\tau$, the network's initial logits on its $N$ training problems;
(ii)~fix the feature-coupling signs $s_i=\mathrm{sign}\,z_i(0)$ and start the shared parameters at $v=w=0$;
(iii)~set $\lambda^2=k_u/k_d'$ and $\mu^2=k_s/k_d'$, so that $z_i=u_i+\lambda v+\mu s_iw$;
(iv)~train with the settings of the network run: minibatches of $B=\min(16,N)$ problems, $K=3$ samples, majority vote, REINFORCE, plain SGD and 100 epochs, with step $0.125B/(1+\lambda^2+\mu^2)$, so that a visit moves a problem's own logit by the same amount as the network's step $0.125B/k_d$;
(v)~repeat with 24 seeds, and take the fraction of seeds that end with $\Phi>0.5$ as the predicted probability of collapse.
The three predictors use the same kernel ratios and differ in how much of the training dynamics they keep. The surrogate keeps the heterogeneous initial logits, the finite training horizon and the sampling noise, so the seed $\delta$ of the shared mode arises from the same sources as in the network. The kernel mean-field keeps the logits and the horizon but drops the noise, so only the initial imbalance of the logits seeds the shared mode. The bare threshold keeps only the initial confidence profile, through $G$, and takes $\Lambda_c$ from the toy model.

\subsection{Marginal-entropy remedy}
\label{app:remedy}
The bonus $\beta H(\bar p)$ adds $\beta\ln\frac{1-\bar p}{\bar p}\,p_i(1-p_i)$ to the gradient on $z_i$.
For a common shift $w$, $\bar p-\tfrac12\approx\langle p(1-p)\rangle w$ and $\ln\frac{1-\bar p}{\bar p}\approx-4(\bar p-\tfrac12)$.
Averaging over problems shows that the shared-mode growth rate becomes $\eta\lambda^2(\langle F'\rangle-4\beta\langle p(1-p)\rangle^2)$, which gives \cref{eq:beta}.
At $z=0$, $F'(0)=1/8$ and $p(1-p)=1/4$, so $\beta_c=1/2$, the maximum over $a$.

%% file: tab_meanfield.tex
\begin{table}[h]
\centering\small
\caption{Deterministic mean-field critical amplification $X_cG(a)$ (with $X=\lambda^2N$) against the closed-form prediction $\ln(1/\delta)$.}
\label{tab:mf}
\begin{tabular}{lcccc}
\toprule
$a$ & $\delta=10^{-2}$ & $\delta=10^{-3}$ & $\delta=10^{-4}$ & $\delta=10^{-6}$ \\
\midrule
0.1 & 2.25 & 4.50 & 6.77 & 11.32 \\
0.3 & 3.23 & 5.47 & 7.72 & 12.26 \\
0.5 & 3.60 & 5.83 & 8.07 & 12.59 \\
0.8 & 3.75 & 5.96 & 8.19 & 12.70 \\
1.1 & 3.50 & 5.69 & 7.92 & 12.41 \\
1.25 & 2.99 & 5.16 & 7.38 & 11.85 \\
\midrule
$\ln(1/\delta)$ & 4.61 & 6.91 & 9.21 & 13.82 \\
\bottomrule
\end{tabular}
\end{table}

%% file: sec_appendix_robust.tex
\section{Robustness: GRPO, Group Size, Many Answers, Adam}
\label{sec:robust}

All discussions in the main text use one training setup: REINFORCE with $K=3$, binary answers and plain SGD. Here we vary each of these choices, still in the toy model, to check that the law of \cref{sec:theory} is not an artifact of the defaults.

\begin{figure}[t]
\centering
\includegraphics[width=\textwidth]{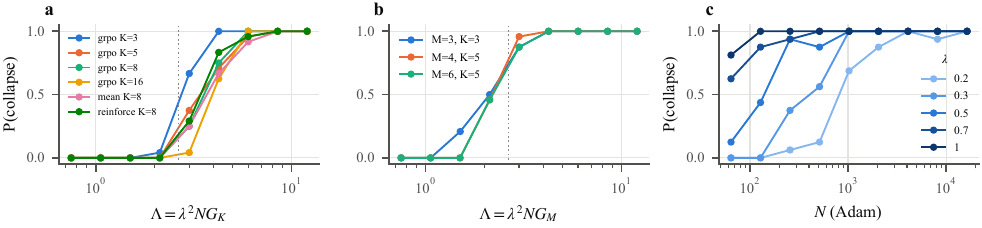}
\caption{\textbf{Robustness.} (a)~GRPO and REINFORCE for $K\in\{3,5,8,16\}$ against $\Lambda=\lambda^2NG_K$, where $G_K$ (\cref{sec:race}) is computed from the exact $K$-sample drift $F_K$ (\cref{sec:single}). (b)~$M$-answer softmax policies against $\Lambda=\lambda^2NG_M$. (c)~Adam: probability of collapse against $N$ for several couplings.}
\label{fig:robust}
\end{figure}

\begin{table}[h]
\caption{Edge of the unstable window $z^\star_K$ (where $F_K'=0$) and the amplification factor $G_K(0.5)=\ln[F^\star_K/F_K(0.5)]$ for different estimators and group sizes $K$ (binary answers). Computed exactly by enumeration.}
\label{tab:window}
\centering\small
\begin{tabular}{lccccc}
\toprule
 & $K=3$ & $4$ & $5$ & $8$ & $16$ \\
\midrule
\multicolumn{6}{l}{\emph{REINFORCE:} $z^\star_K$} \\
 & 1.317 & 1.274 & 1.187 & 1.088 & 0.894 \\
\multicolumn{6}{l}{\emph{GRPO (std-normalized):} $z^\star_K$} \\
 & 1.317 & 1.317 & 1.213 & 1.172 & 1.006 \\
\midrule
\multicolumn{6}{l}{\emph{REINFORCE:} $G_K(0.5)$} \\
 & 0.514 & 0.484 & 0.426 & 0.353 & 0.202 \\
\multicolumn{6}{l}{\emph{GRPO:} $G_K(0.5)$} \\
 & 0.514 & 0.514 & 0.443 & 0.408 & 0.258 \\
\bottomrule
\end{tabular}
\end{table}

\paragraph{GRPO and group size.}
For $K=3$, GRPO's drift is exactly $\sqrt2F$ (\cref{sec:single}). The constant factor cancels in $G$, so $G_3(0.5)=0.514$ for both estimators (\cref{tab:window}), and the theory predicts the same $N_c$.
The simulations agree: the logistic fit to the GRPO $K=3$ curve in \cref{fig:robust}a gives $N_c=548$, against $N_c=536$ for REINFORCE at the same $\lambda=0.1$ and $a=0.5$ (\cref{fig:scaling}c).
Larger groups have smaller $G_K$ (\cref{tab:window}), so the theory predicts larger $N_c$. The fitted values rise to $793$, $876$ and $1557$ for $K=5,8,16$ (\cref{fig:robust}a), in the order predicted by $G_K$.
Computing $\Lambda_c=\lambda^2N_cG_K$ from these fits, the critical amplification drifts mildly upward with $K$ ($\Lambda_c=2.8\to4.0$), consistent with the smaller gradient noise and hence smaller seed $\delta$ of larger groups.
The mean baseline and REINFORCE at $K=8$, also shown in \cref{fig:robust}a, give $\Lambda_c=3.8$ and $3.5$, within this band.

\paragraph{Many answers.}
Next we replace binary answers by $M$-answer softmax policies, as in the theory of \cref{sec:general}.
With $M\in\{3,4,6\}$ answers, softmax logits and a shared bias \emph{vector}, the collapsed state has every problem giving the same answer, and accuracy drops to $1/M$.
Using the eigenvalue-based $G_M$ of \cref{sec:general}, computed by exact enumeration (\cref{app:multi}), the transitions fall at $\Lambda_c=2.1$--$2.2$ (\cref{fig:robust}b), slightly below the binary value. This is consistent with $M-1$ noisy shared directions seeding the instability.

\paragraph{Adam.}
The last change is the optimizer. Unlike the changes above, it alters the rates in the theory itself, so we first rederive the scaling.
Adam \citep{kingma2015adam} normalizes each coordinate by its own gradient scale, which changes both rates in the race.
The shared parameter sees a gradient dominated by noise of scale $\lambda\sigma/\sqrt B$, so its signal-to-noise-normalized drift makes the shared mode grow at a rate $\propto\lambda\langle F'\rangle$ rather than $\lambda^2\langle F'\rangle$.
A problem-specific coordinate is updated only on a fraction $B/N$ of steps, so its normalized step grows like $\sqrt{N/B}$ per epoch.
The ratio gives $\Lambda_{\mathrm{Adam}}\propto\lambda\sqrt N\,G$, and therefore the \emph{same} scaling $N_c\propto\lambda^{-2}$ with a different constant.
Simulations agree (\cref{fig:robust}c). For $\lambda\in\{0.2,0.3,0.5\}$ we find $N_c=899,378,136$, so $\lambda^2N_c=34$--$36$ is constant (fitted slope $-2.06$). With our hyperparameters this constant is about $9\times$ the SGD value at the same coupling ($\lambda^2N_c\approx4.0$ at $\lambda=0.2$, \cref{fig:scaling}c), so Adam is more tolerant, but the law is unchanged.
(For $\lambda\ge0.7$, $N_c$ falls below the smallest size we simulate.)

%% file: tab_conditions.tex
\begin{table}[h]
\centering\small
\caption{The 22 Transformer conditions of \cref{sec:nn}. The kernel ratios and the amplification $G$ are measured on the initial network before any RL step. The base column gives the number of supervised warm-up steps and the resulting held-out accuracy. The last column counts collapsed runs out of 42 (7 values of $N$ and 6 seeds).}
\label{tab:conditions}
\begin{tabular}{llcccccc}
\toprule
model & base (acc.) & $\tau$ & $\rho_b$ & $k_u/k_d'$ & $k_s/k_d'$ & $G$ & collapsed \\
\midrule
plain & 90 (82\%) & 1 & -- & 0.702 & 0.051 & 0.025 & 29/42 \\
plain & 90 (82\%) & 2 & -- & 0.702 & 0.051 & 0.315 & 34/42 \\
plain & 90 (82\%) & 3 & -- & 0.702 & 0.051 & 0.622 & 37/42 \\
\addlinespace
plain & 120 (89\%) & 1 & -- & 0.404 & 0.058 & 0.000 & 17/42 \\
plain & 120 (89\%) & 2 & -- & 0.404 & 0.058 & 0.028 & 35/42 \\
plain & 120 (89\%) & 3 & -- & 0.404 & 0.058 & 0.186 & 38/42 \\
\addlinespace
antisymmetric & 10 (84\%) & 1 & 0 & 0.001 & 0.156 & 0.532 & 0/42 \\
antisymmetric & 10 (84\%) & 1 & 0.3 & 0.349 & 0.158 & 0.532 & 32/42 \\
antisymmetric & 10 (84\%) & 1 & 1 & 1.164 & 0.161 & 0.532 & 42/42 \\
antisymmetric & 10 (84\%) & 1 & 3 & 3.521 & 0.171 & 0.532 & 42/42 \\
antisymmetric & 10 (84\%) & 2 & 0 & 0.001 & 0.156 & 1.125 & 0/42 \\
antisymmetric & 10 (84\%) & 2 & 0.3 & 0.349 & 0.158 & 1.125 & 41/42 \\
antisymmetric & 10 (84\%) & 2 & 1 & 1.164 & 0.161 & 1.125 & 42/42 \\
antisymmetric & 10 (84\%) & 2 & 3 & 3.521 & 0.171 & 1.125 & 42/42 \\
\addlinespace
antisymmetric & 20 (92\%) & 1 & 0 & 0.002 & 0.446 & 0.177 & 0/42 \\
antisymmetric & 20 (92\%) & 1 & 0.3 & 0.436 & 0.446 & 0.177 & 22/42 \\
antisymmetric & 20 (92\%) & 1 & 1 & 1.451 & 0.447 & 0.177 & 41/42 \\
antisymmetric & 20 (92\%) & 1 & 3 & 4.361 & 0.451 & 0.177 & 42/42 \\
antisymmetric & 20 (92\%) & 2 & 0 & 0.002 & 0.446 & 0.520 & 0/42 \\
antisymmetric & 20 (92\%) & 2 & 0.3 & 0.436 & 0.446 & 0.520 & 37/42 \\
antisymmetric & 20 (92\%) & 2 & 1 & 1.451 & 0.447 & 0.520 & 42/42 \\
antisymmetric & 20 (92\%) & 2 & 3 & 4.361 & 0.451 & 0.520 & 42/42 \\
\bottomrule
\end{tabular}
\end{table}

%% file: tab_nn.tex
\begin{table}[h]
\centering\small
\caption{Kernel drift during label-free RL (one seed per case). Ratios are reported only before collapse ($\Phi<0.5$): once all answers agree, $s\propto\mathbf 1$ and the decomposition degenerates. $k_d(t)/k_d(0)$ is the growth of the self-kernel, i.e.\ of the effective step size.}
\label{tab:drift}
\begin{tabular}{lrcccc}
\toprule
case & epoch & $k_u/k_d'$ & $k_s/k_d'$ & $k_d(t)/k_d(0)$ & $\Phi$ \\
\midrule
antisym 10, $\tau$=1, $\rho_b$=0.3, $N$=1024 & 0 & 0.345 & 0.152 & 1.0 & 0.02 \\
 & 1 & 0.009 & 0.222 & 28.1 & 0.28 \\
 & 4 & 0.002 & 0.074 & 101.9 & 0.14 \\
\addlinespace
antisym 20, $\tau$=1, $\rho_b$=0.3, $N$=2048 & 0 & 0.429 & 0.423 & 1.0 & 0.02 \\
 & 1 & 0.219 & 0.066 & 1.4 & 0.21 \\
 & 4 & 0.121 & 0.051 & 2.2 & 0.33 \\
 & 8 & 0.082 & 0.043 & 3.3 & 0.32 \\
\addlinespace
plain 90, $\tau$=1, $N$=512 & 0 & 0.700 & 0.050 & 1.0 & 0.04 \\
 & 1 & 0.683 & 0.092 & 1.1 & 0.25 \\
\addlinespace
plain 120, $\tau$=1, $N$=1024 & 0 & 0.405 & 0.059 & 1.0 & 0.14 \\
 & 1 & 0.384 & 0.056 & 1.2 & 0.04 \\
 & 4 & 0.304 & 0.044 & 1.6 & 0.16 \\
 & 8 & 0.254 & 0.037 & 2.2 & 0.12 \\
\addlinespace
antisym 10, $\tau$=1, $\rho_b$=0, $N$=1024 & 0 & 0.000 & 0.153 & 1.0 & 0.02 \\
 & 1 & 0.000 & 0.215 & 46.3 & 0.03 \\
 & 4 & 0.001 & 0.133 & 245.2 & 0.02 \\
 & 8 & -0.001 & 0.236 & 186.1 & 0.05 \\
\addlinespace
\bottomrule
\end{tabular}
\end{table}